\documentclass{ecai}

\usepackage{latexsym}
\usepackage{amssymb}
\usepackage{amsmath}
\usepackage{amsthm}
\usepackage{booktabs}
\usepackage{enumitem}
\usepackage{graphicx}
\usepackage{color}
\usepackage{caption}

\usepackage{anyfontsize}
\usepackage{xcolor}
\usepackage{mathtools}
\usepackage{enumitem}

\usepackage{booktabs}
\usepackage{multirow}
\usepackage{bm}
\usepackage{adjustbox}

\usepackage{flushend}
\usepackage{placeins}
\usepackage{appendix}

\newtheorem{theorem}{Theorem}

\newtheorem{example}{Example}
\newtheorem{remark}{Remark}

\newcommand{\BibTeX}{B\kern-.05em{\sc i\kern-.025em b}\kern-.08em\TeX}

\DeclareMathOperator{\E}{\mathbb{E}}
\newcommand{\I}{\mathbb{I}}
\newcommand{\R}{\mathbb{R}}
\newcommand*\dif{\mathop{}\!\mathrm{d}}

\usepackage{xr}
\makeatletter
    \newcommand*{\addFileDependency}[1]{
    \typeout{(#1)}
    \@addtofilelist{#1}
    \IfFileExists{#1}{}{\typeout{No file #1.}}
    }
\makeatother

\newcommand*{\myexternaldocument}[1]{%
    \externaldocument{#1}%
    \addFileDependency{#1.tex}%
    \addFileDependency{#1.aux}%
}

\myexternaldocument{main-ecai}
\myexternaldocument{main-appendices}

\begin{document}


\begin{frontmatter}


    \paperid{1980 } 


    \title{Augmented prediction of a true class for Positive Unlabeled data under selection bias}


    \author[A,B]{\fnms{Jan}~\snm{Mielniczuk}\orcid{0000-0003-2621-2303}\thanks{Corresponding Author. Email: jan.mielniczuk@ipipan.waw.pl.}\footnote{Equal contribution.}}
    \author[A]{\fnms{Adam}~\snm{Wawrzeńczyk}\orcid{0000-0002-6202-7829}\footnotemark}

    \address[A]{Institute of Computer Science, Polish Academy of Sciences}
    \address[B]{Faculty of Mathematics and Information Science, Warsaw University of Technology}


    \begin{abstract}

        We introduce a new observational setting for Positive Unlabeled (PU) data where the observations at prediction time are also labeled.
        This occurs commonly in practice -- we argue that the additional information is important for prediction, and call this task "augmented PU prediction".
        We allow for labeling to be feature dependent.
        In such scenario, Bayes classifier and its risk is established and compared with a risk of a classifier which for unlabeled data is based only on predictors. We introduce several variants of the empirical Bayes rule in such scenario and investigate their performance.
        We emphasise dangers (and ease) of applying classical classification rule in the augmented PU scenario -- due to no preexisting studies, an unaware researcher is prone to skewing the obtained predictions.
        We conclude that the variant based on recently proposed variational autoencoder designed for PU scenario works on par or better than other considered variants and yields advantage over feature-only based methods in terms of accuracy for unlabeled samples.
 
    \end{abstract}

\end{frontmatter}


\section{Introduction}
\label{section:intro}

    We consider Positive Unlabeled (PU) learning, that is a binary classification task in which information about class indicators is only partially observable. More specifically, in the PU scenario some observations from a positive class are assigned labels, whereas the remaining observations from this class, as well as all negative observations, are unlabeled. PU data is collected in many practical situations, usually when obtaining reliable negatives is difficult or costly. Under such scenario the most common Machine Learning task is construction of a classification rule based \textit{only} on predictors, which will assign a new observation to a positive or a negative class. In genetics, one can find some genes influencing a particular disease via costly experiments, but it cannot be assumed that the other genes in GenBank are not relevant for this disease~\citep{PU-genetics}. Other typical uses include e.g. ecology~\citep{Ward2009}, survey analysis~\citep{BekkerDavis2020, Sechidis2017}, recommendation systems~\citep{Schnabel, XMLC, Chen2021,Naumov2019} and fraud detection~\citep{Luo2018}. PU framework can be considered as a special case of data with noisy or partial labels \citep{Menon2018, Cannings2020, Liu2014,Hull2014, Cabannes2020}. However, there are PU learning problems where the data partial observability can be slightly loosened -- in many applications we would like to perform classification on {\it new} PU observations for which along with predictors, \textit{the labeling status is given}.    
    
    Such situations commonly happen. A typical example is occurrence of hypertension. People who check their blood pressure regularly and when it is abnormal report that to a doctor, are treated for hypertension. In such a case positive labels are assigned to them. However, the remaining (unlabeled) group consists of those who have abnormal blood pressure level but do not contact a doctor, and those who are healthy. Another example is reporting episodes of certain illness (i.e. migraine) using dedicated software, see e.g. \citep{Park2016}. Here, some patients who experience such episodes, fail to report them and thus they can not be distinguished from patients who do not have them, whence both groups fall into unlabeled category. Note that in the examples labeling may depend on characteristics of the patients: e.g. in the first one the better educated and thus more aware of consequences of untreated hypertension are more likely to consult a doctor, in the second, the age, influencing dexterity of using a dedicated smartphone application, may be an important factor. This is called selection bias (or instance-dependent labeling) and correspond to the fact that the distribution of selected (i.e. labeled) observations is different from that of a positive class. 

    Note that in the considered examples labeling of new observations occur naturally.
    In the first example above, in a new batch of patients, for those who fail to report hypertension, one would like to detect those who are likely to be positive. In the "migraine" example it is of interest to detect patients who likely have failed to report migraine episodes, in order to contact them. Of course, in such a case, assignment is an issue for unlabeled observations only, as for the labeled ones we know for sure that they belong to a positive class. 
    For the sake of distinguishing such task from the usual classification based on predictors alone, we will call this problem prediction for augmented PU observations or, in short, {\it augmented PU prediction}.To the best of our knowledge the paper is the first approach discussing this problem in the literature.
    
    In the paper we establish the form of Bayes selection rule for detection of positive observations among unlabeled ones and show that it is more conservative than Bayes classification rule based solely on predictors. The fact that we are less likely to classify items to a positive class {\it when they are unlabeled} is understandable when one realises that unlabeled class contains relatively {\it less} positive observations than the general population.

    We calculate the Bayes risk for such scenario and bound the excess risk for an classification based solely on predictors, what sheds light on advantage of using labeling information. Also we introduce empirical Bayes classifiers taking advantage of recent proposals for posterior probability estimators in this context. We show that the variant based on variational autoencoder designed for PU data works promisingly when accuracy relative to unlabeled data is considered as an evaluation metric.
    
\section{Preliminaries}

    In the PU scenario, one considers a random vector $(X, Y, S)$ with a distribution $P_{X, Y, S}$ such that $X \in \R^p$ and $Y,S$ are binary with values 0 or 1. $Y$ is a class indicator, with $Y = 1$ denoting a positive class and $Y = 0$, a negative one, whereas $S = 1$ and $S = 0$ mean that observation is either labeled or unlabeled, respectively. The considered setting stipulates that only some positive observations are labeled, whereas the remaining positive observations and negative ones are unlabeled. We adopt Selected At Random (SAR) assumption, which states that probability of labeling positive observation depends on observed values of   predictors corresponding to it. Note that it is less stringent, and, as mentioned in the Introduction, more realistic than assumption that labeling is random but independent of an observation's features (Selected Completely At Random or SCAR assumption). 
    
    In single training-sample scenario adopted here it is also assumed that random iid vectors $(X_i, Y_i, S_i), i = 1, \ldots, n$ are generated according to $P_{X, Y, S}$, but the observable data is ${\cal T}=\{(X_i, S_i), i=1, \ldots, n\}$. This is in contrast to case-control case when it is assumed that two $X$ samples are available, one pertaining to the positive class (i.e. sampled from $P_{X|Y = 1}$) and the second corresponding to the general population (that is, sampled from $P_X$).
    
    The basic numerical quantities partially describing distribution $P_{X, Y, S}$ are (unobservable) posterior probability of a positive class $y(x) = P(Y = 1|X = x)$ and (observable) posterior probability of being labeled $s(x) = P(S = 1|X = x)$. Note that since the considered mechanism ensures that $P(S = 1|Y = 0, X = x) = 0$, the law of total probability implies the following relation between them
    \begin{equation}
    \label{eq:basic}
        \begin{aligned}
            s(x) 
                & = P(S = 1|Y = 1, X = x) P(Y = 1|X = x) \\
                & := e(x) y(x), 
        \end{aligned}
    \end{equation}
    where $e(x)$, probability of being labeled given that it is positive and $X = x$, is called propensity score. Note that under SAR assumption this is, usually not constant, function of vector of predictors $x$. Risk bounds when $e(x)$ is known are given in \citep{Coudray2023}. 
    Under SCAR $s(x)$ is constant and equals probability of a positive element being labeled $P(S=1|Y=1)$, which will be denoted by $c$.

    We briefly discuss PU research in SAR setting and single-training-sample (or censoring) scenario which gains momentum recently due to its less stringent assumptions on labeling mechanism. The main approach to model posterior probability of positive outcome $Y=1$ and propensity score as parametric functions. Furthermore, treating $Y$ as a hidden random variable one employs Expectation-Maximization (EM) algorithm to estimate them \citep{Gong2021}. It is also possible to alternately optimize estimates of their Fisher consistent expressions \citep{BekkerRobberechtsDavis2019}. Another approach avoids estimation of propensity function and uses Empirical Risk Minimization method along with modelling of posterior by deep NN to find a solution \citep{NaVAE, VAEPUCC}. Other methods use additional assumptions such as co-monotonicity of posterior probabilities for $Y$ and for $S$ or some form of functional relation between posterior and propensity score \citep{Gerych2022}. We will use modified version of variational autoencoder proposed in \cite{VAEPUCC} to solve augmented PU prediction problem discussed here. We also mention case-control scenario, in which a selection bias is recently incorporated \citep{Kato2019, AAAI2023}. 

\section{Augmented PU prediction method and its properties}

    We consider now augmented prediction for PU observations (augmented PU prediction) scenario when a new observation $(X, S)$ is given and we want to predict the corresponding value of $Y$. Obviously, when $S = 1$ under assumed scenario we have $Y = 1$ and thus we need to consider only the case $S = 0$. We introduce the following prediction rule
    \begin{equation}
    \label{eq:bayes_PU}
        d_{B}^{PU}(x,s) =
        \begin{cases}
            1, & \text{if}\ s = 1 \\
            \begin{cases}
                1, & \text{if}\ y(x) > \frac{1 + s(x)}{2}\\
                0, & \text{otherwise,}
            \end{cases} & \text{if}\ s=0 \\
        \end{cases}
    \end{equation}
    where $y(x)$ is posterior probability of positive class defined above (\ref{eq:basic}).
    We will investigate the loss of efficiency when label $S$, which carries information about $Y$, is not available for classification. To this end we consider Bayes rule $d_B(x)$ based solely on $x$:
    \begin{equation}
    \label{eq:bayes_x}
        d_{B}(x) =
        \begin{cases}
            1, & \text{if}\ y(x) > \frac{1}{2} \\
            0, & \text{otherwise.}
        \end{cases} 
    \end{equation}
    Directly from the above definitions we have that $d_B^{PU}(X,S)$ is more conservative on class $S=0$ than $d_B(X)$ i.e. it less likely assigns objects to the positive class:
    \[ 
        P(d_B^{PU}(X, S) = 1 | S = 0) \leq P(d_B(X) = 1 | S = 0).
    \]
    Below we show that the rule $d_{B}^{PU}$ is optimal for 0-1 loss and calculate its risk and the excess risk of $d_B(x)$. The fact that the optimal rule is less likely to assign positive class to unlabeled observations than $d_B$ is due to the fact that positive observations occur less frequently among unlabeled ones than in general population. Note that as $d_{B}^{PU}$ is more conservative, classification changes might occur for both positive and negative examples -- though in the expectation, the precision gain should outweigh the lost recall. Also, it has practical consequences for recommendations on the thresholds applied for classification in the follow-up studies involving PU data; see Section \ref{Numerical}. The introduced approach is based on a simple observation that the considered problem can be regarded as a problem of determining the Bayes risk in the case when the vector of predictors is augmented by an additional predictor $S$. This also motivates the name of the problem. 
    We let 
    \begin{equation}
    \label{def:tilde y}
     \tilde{y}(x, s) = P(Y = 1|(X, S) = (x, s)) 
    \end{equation}
    be posterior probability of $Y = 1$ given the augmented vector of predictors. We also define the excess risk (or regret) of any augmented PU prediction rule $d(x,s)$ as (see e.g. \cite{menon15}):
    \[\Delta(d) = P\left(d(X,S) \neq Y \right) - P\left(d_{B}^{PU}(X, S) \neq Y \right).\]
    \begin{theorem}
        \label{theorem:m}
        \begin{enumerate}[label=(\roman*)]
            \item $d_{B}^{PU}(X, S)$ defined in (\ref{eq:bayes_PU}) is the Bayes rule for $Y$ under $P_{X, Y, S}$ i.e. it is the classification rule yielding the smallest misclassification error $P(d(X, S) \neq Y)$. Moreover, $d_{B}^{PU}(X, 0)$ is the Bayes rule for $Y$ under $P_{X, Y|S = 0}$ yielding the smallest classification error $P(d(X) \neq Y|S = 0)$.\\
            \item Define $w(x) = 1 + s(x) - 2 y(x)$. Then Bayes risk of $d_{B}^{PU}(x, s)$ equals
            \begin{equation}
            \label{eq:B_risk}
                \begin{aligned}
                    L^*_{PU}
                        & = \frac{1}{2} \Big(P(S = 0) - \E_{X, S = 0} \left|2 \tilde{y}(X, 0) - 1 \right| \Big) \\
                        & = \frac{1}{2} \Big(P(S = 0) - \E_X \left|w(X) \right| \Big)
                \end{aligned}
            \end{equation}
            \item We have for excess risk of $d_B(x)$:
            \begin{equation}
            \label{eq:excess}
                \E_X \left(s(X) \I \left\{y(X) < \frac{1}{2} \right\} \right)
                    \leq \Delta(d_B)
                    \leq P(S = 1). 
            \end{equation}
            The inequalities above are tight when $P\left(y(X) < \frac{1}{2}\right) = 1$.
            \item Odds ratio $OR(x)$ for odds of $Y = 1$ in class $\{S = 0\}$ and odds of $Y = 1$ in a general population equals
            \begin{equation}
            \label{eq:OR}
                \begin{aligned}
                    OR(x)
                        & = \frac{P(Y = 1|S = 0, X = x)}{P(Y = 0|S = 0, X = x)}
                        \, \bigg/ \, \frac{P(Y = 1|X = x)}{P(Y = 0|X = x)} \\
                        & = 1 - e(x).
                \end{aligned}
            \end{equation}
        \end{enumerate}
    \end{theorem}

    \begin{proof}
        \begin{enumerate}[label=(\roman*)]
            \item As we have
                \begin{equation*}
                    \begin{aligned}
                        P\Big(d_{B}^{PU}(X, S) \neq Y \Big)
                            & = P\Big(d_{B}^{PU}(X, S) \neq Y \big| S = 1 \Big) P(S = 1) \\
                            & + P\Big(d_{B}^{PU}(X, S) \neq Y \big| S = 0 \Big) P(S = 0)
                    \end{aligned}
                \end{equation*}
                and the first term on RHS equals 0, it is enough to prove that the second and the third line in (\ref{eq:bayes_PU}) define Bayes rule on the strata $\{S = 0\}$. The Bayes rule for this problem is given by assigning $Y = 1$ when the following condition holds:
                \[ 
                    \frac{P(Y = 1|S = 0, X = x)}{P(Y = 0|S = 0,X = x)} > 1.
                \]
                Denoting by $f(x)$ either the density of $X$ or its probability mass function at $x$ we have, inverting conditional probabilities, that that the ratio above equals
                \begin{equation}
                \label{eq:ratio}
                    \begin{aligned}
                        \frac{P(S = 0, Y = 1, X = x)}{P(S = 0, Y = 0, X = x)}
                                & = \frac{f(x) (y(x) - s(x))}{f(x) (1 - y(x))} \\
                                & = \frac{y(x) - s(x)}{1 - y(x)}.
                    \end{aligned}
                \end{equation}

                Then it is enough to note that
                \begin{equation}
                \label{eq:equiv}
                    \frac{y(x) - s(x)}{1 - y(x)} > 1 
                    \,\,\,\,\, \equiv \,\,\,\,\, 
                    y(x) > \frac{1 + s(x)}{2}.
                \end{equation}
            \item As $d_{B}^{PU}(x,s)$ is Bayes classifier its risk equals 
            \begin{equation}
            \label{eq:min_formula}
                L^*_{PU} 
                    = \E_{X, S} \min{\big(\tilde{y}(X, S), 1 - \tilde{y}(X, S) \big)},
            \end{equation}
            where $\tilde{y}(x, s)$ is defined in (\ref{def:tilde y}).
            This is easily justified by noting that if $\tilde{y}(x, s) > \frac{1}{2}$ and thus $(x, s)$ is assigned to a positive class by the Bayes classifier, it commits an error with probability $1 - \tilde{y}(x, s) = \min{\left(\tilde{y}(x, s), 1 - \tilde{y}(x, s) \right)}$. Moreover, we have that $\tilde{y}(x, 1) = 1$ and reasoning as in (\ref{eq:ratio}) we obtain
            \begin{equation*}
                \begin{aligned}
                    \tilde{y}(x, 0)
                        & = P(Y = 1 | (X, S) = (x, 0)) \\
                        & = \big( y(x) - s(x) \big) \,/\, \big( 1 - s(x) \big).                    
                \end{aligned}
            \end{equation*}
            In view of $\min(a, b) = {(a+b - |b-a|)} / {2}$ we have $\min(a, 1 - a)=(1 -|2a - 1|) / 2$ and whence (\ref{eq:min_formula}) implies that
            \begin{equation}
            \label{eq:PUrisk}
                \begin{aligned}
                    L^*_{PU} 
                        & = \frac{1}{2} - \frac{1}{2} \E_{X, S} \Big|2\tilde{y}(X, S) - 1 \Big| \\
                        & = \frac{1}{2} 
                            - \frac{1}{2} \E_{X, S = 1} \Big|2\tilde{y}(X, 1) - 1 \Big| \\
                            & \phantom{= \frac{1}{2} \ }
                             - \frac{1}{2} \E_{X, S = 0} \Big|2\tilde{y}(X, 0) - 1 \Big| \\
                        & = \frac{1}{2}
                            - \frac{1}{2} P(S = 1) 
                            - \frac{1}{2} \E_{X, S = 0} \Big|2\tilde{y}(X, 0) - 1 \Big|.
                \end{aligned}
            \end{equation}
           Thus we established the first equality in (\ref{eq:B_risk}). Noting that
            \begin{equation*}
                \begin{aligned}
                    & \E_{X, S = 0} \big| 2 \tilde{y}(X, 0) - 1 \big| \\
                        & = \int \frac{\big| 2 y(x) - s(x) - 1 \big|}{1 - s(x)} f(x) (1 - s(x)) \, \dif x \\
                        & = \E_X |w(X)| 
                \end{aligned}
            \end{equation*}
            we establish the second one. We note that from the proof above it follows that $d_{B}^{PU}(x, 0)$ is the Bayes classifier on the strata $\{S = 0\}$ and its Bayes risk $ L^{*0}_{PU}$ equals 
            \begin{equation}
            \label{eq:risk_S=0}
                \begin{aligned}
                L^{*0}_{PU}
                    = \frac{L^*_{PU}}{P(S = 0)}
                    & = \frac{1}{2} - \frac{1}{2} \E_{X|S = 0} \big| 2\tilde{y}(X, 0) - 1 \big| \\
                    & = \frac{1}{2} - \frac{\E_X|w(x)|}{P(S = 0)}.
                \end{aligned}
            \end{equation}
            \item Reasoning as above we have
            \[
                L^*
                    = P(d_B(X) \neq Y)
                    = \frac{1}{2} - \frac{1}{2} \E_X \big| 2 y(X) - 1 \big|
            \] 
            and in view of (\ref{eq:PUrisk}) we obtain
            \begin{equation*}
                \begin{aligned}
                    L^* - L_{PU}^*
                        & = \frac{1}{2} P(S = 1) \\
                        & + \frac{1}{2} \E_X \Big\{ 
                            \big| 2 y(X) - s(X) - 1 \big| - \big| 2 y(X) - 1 \big|
                        \Big\}.
                \end{aligned}
            \end{equation*}
            RHS of (\ref{eq:excess}) is obtained by using triangle inequality $\big| 2 y(X) - s(X) - 1 \big| - \big| 2 y(X) - 1 \big| \leq s(x)$. To prove LHS of (\ref{eq:excess}) we note that we have the following refinement of triangle inequality for $b \geq 0$
            \[
                |a - b| \geq |a| - b + 2 b \times \I\{a < 0\}
            \]
            Applying this to $a := 2 y(X) - 1$ and $b := s(X)$ we have that 
            \[
                \big| 2 y(X) - s(X) - 1 \big| \geq \big| 2 y(X) - 1 \big| - s(x) + 2 s(X) \I\left\{y(X) < \frac{1}{2}\right\}
            \] 
            and this implies the conclusion. Note that the lower bound equals the upper bound when for all $x$ we have $y(x) < \frac{1}{2}$. In this case we note that $P \big( d_B(X) \neq Y \big) = P(Y = 1)$ whereas ${P \big( d_B^{PU}(X, S) \neq Y \big)} = {P(S = 0, Y = 1)}$ and the excess risk is thus $P(Y = 1) - P(S = 0, Y = 1) = P(S = 1).$ The result in (iii) is intuitive: $d_B^{PU} $ does not err on $S=1$, whereas $d_B$ commits an error on this stratum if $y(x)<1/2$.
            \item This follows by noting that in view of above derivations
            \[
                OR(x) 
                    = \frac{y(x) - s(x)}{1 - y(x)} \, \Big/ \,\, \frac{y(x)}{1 - y(x)} 
                    = \frac{y(x) - s(x)}{y(x)}
                    = 1 - e(x).
            \]
        \end{enumerate}
    \end{proof}

    \begin{remark}
        \begin{enumerate}[label=(\roman*)]
            \item The threshold in (\ref{eq:bayes_PU}) can be expressed as
            \[
                y(x)>\frac{1+s(x)}{2}\equiv y(x)>\frac{1}{2-e(x)}.
            \]
              When $e(x)$ is large, then unlabeled element is less likely to be positive and the threshold becomes larger.
              \item We note that when labeling is independent of an object in a positive class (SCAR assumption) and thus propensity score $e(x)\equiv c$, we have (cf. (i)):\\
            \[
                d_B^{PU}(x, 0) = 1\,\, \iff \,\, y(x) > \frac{1}{2 - c}.
            \]
            For situation of complete lack of labeling ($c=0$) unlabeled class is distributed according to $P_X$ and $ d_B^{PU}(x, 0)$ coincides with $d_B(x)$ in agreement with the last inequality. Note that since under SCAR positive observations are labeled or not, regardless of the predictors' values, the threshold $(2 - c)^{-1}$ above is due \textit{solely} to the changed proportion of positives among unlabeled ones compared with the general population.
            \item The result can be generalised to strictly proper composite losses $\ell(s, y)$ such that corresponding Bayes classification function equals $\phi(OR(x))$ and $\phi$ is strictly increasing as e.g. for logistic loss $\ell_{logistic}(s, y) = \log(1 + \exp(-sy))$ for which $\phi(s) = s$. Then the Bayes rule on the class $S = 0$ assigns $x$ to class $Y = 1$ when $y(x) > (\phi^{-1}(1) + s(x)) / (1 + \phi^{-1}(1))$. In particular it is equal to $d_B^{PU}(x)$ for logistic loss. 
        \end{enumerate}
    \end{remark} 

    Below we calculate excess risk in (\ref{eq:excess}) for a specific model.
    \begin{example}
        Let $y(x) = \Phi(x), X \sim N(0, 1)$, and $x \in \R$ (univariate probit model with standard normal predictor), and let propensity score $e_a(x) = \I \{x > a\}$ i.e. above threshold $a \in \R$ all positive observations are labeled. In this case the excess risk of $d_B(x)$ defined in (\ref{eq:bayes_x}) for $a > 0$ equals (refer to appendix~\ref{appendix:example_1} for full derivation)
        \begin{equation*}
            \begin{aligned}
                & \Delta(d_B)=\E_X \big[\min{\big( y(X), 1 - y(X) \big)} \big] \\
                    & \phantom{= \ } - \E_{X, S} \big[\min{\big( \tilde{y}(X, S), 1 - \tilde{y}(X, S) \big)} \big] \\
                & = \frac{1}{2} - \Phi(a) + \frac{\Phi^2(a)}{2}
                    = \frac{1}{2} \big(\Phi(a) - 1 \big)^2 \geq 0,
            \end{aligned}
        \end{equation*}
        and for $a < 0$ equals $\frac{1}{4} -\frac{\Phi^2(a)}{2}\geq 0$. Note that for $a\to \infty$ excess risk tends to 0 as $P_{X, S = 0}$ approaches $P_X$ in this case and $d_B^{PU}(x, 0)$ tends to $d_B(x)$. For $a\to -\infty$ the excess risk tends to 1/4 (risk of $d_B(x)$) as the risk of $d_B^{PU}(x, s)$ tends to 0. 
    \end{example}

    \begin{example}
        Consider the situation when $y(x) = \sigma(\alpha x)$ and $e(x) = \sigma(\beta x)$ for $x \in \R$ and $ \alpha,\beta\geq 0$.
        Then we have for $\tilde y(x,0)$ defined in (\ref{def:tilde y})
        \begin{equation}
            \begin{aligned}
                \tilde{y}_{\alpha,\beta}(x, 0)
                    & = \frac{y(x) - s(x)}{1 - s(x)}
                        = \frac{\sigma(\alpha x) - \sigma(\alpha x) \sigma(\beta x)}{1 - \sigma(\alpha x) \sigma(\beta x)} \\
                    & = \frac{\frac{1}{\sigma(\beta x)} - 1}{\frac{1}{\sigma(\alpha x) \sigma(\beta x)} - 1}
                        = \frac{1}{1 + e^{-(\alpha - \beta) x} + e^{-\alpha x}}.
            \end{aligned}
        \end{equation}
        The plot of $\tilde{y}_{\alpha, \beta}(x, 0)$ for $\alpha = 1$ and various $\beta$s is shown on Figure \ref{fig:example}. Note that for $\alpha = \beta$ we have $\tilde{y}_{\alpha, \alpha}(x, 0) = (2 + \exp(-\alpha x))^{-1}$ which tends to $\frac{1}{2}$ when $x \to +\infty$, indicating the most difficult situation when $\tilde{y}(x, 0)$ is in a vicinity of $\frac{1}{2}$.
        \begin{figure}
            \centering
            \includegraphics[width=.45\textwidth]{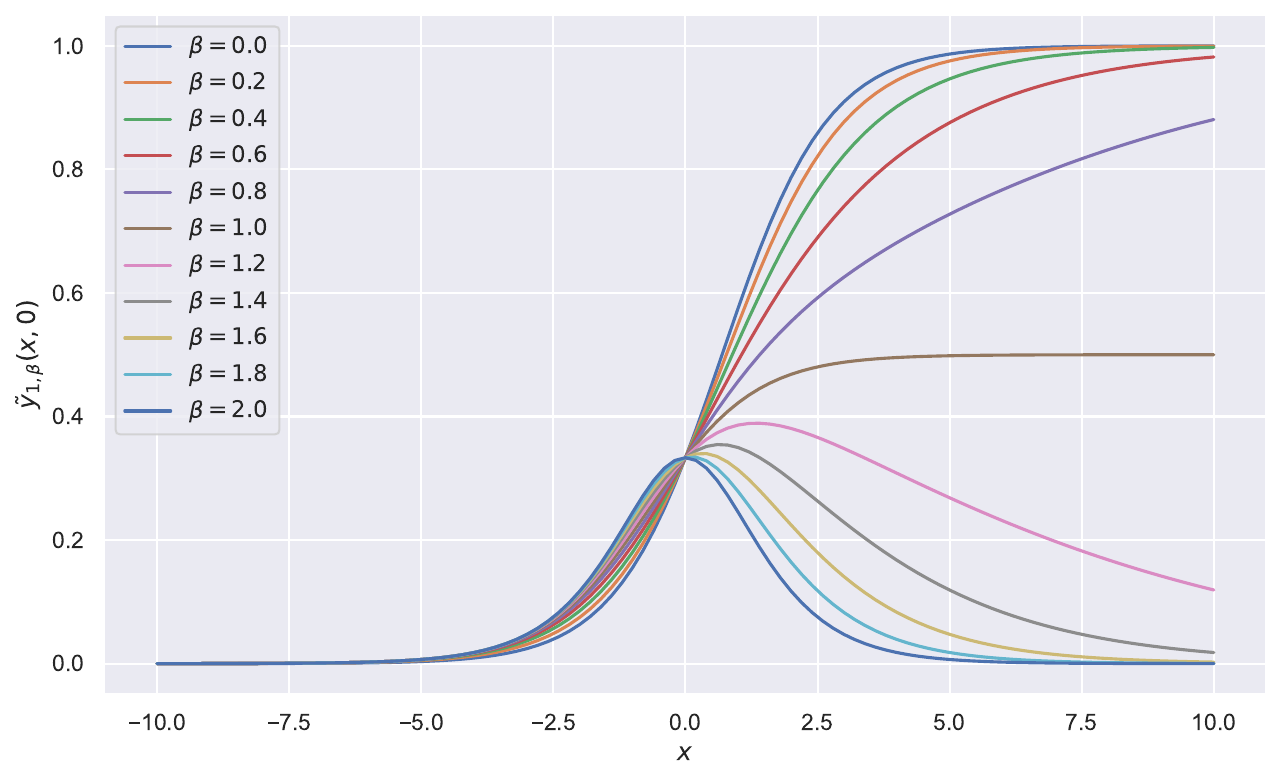}
            \caption{Values of $\tilde{y}_{1, \beta}(x, 0)$ depending on $\beta$.}
            \label{fig:example}
        \end{figure}
    \end{example}

\section{$d_{B}^{PU}$ applications -- VAE-PU-Bayes}
\label{section:vae-pu-bayes}

    The proposed $d_{B}^{PU}$ rule uses are not limited to the direct applications to the augmented PU prediction style data (where the observation label is available for the test data). As a motivational   example we consider first a typical PU problem, with only predictors available at the test time.
    
    VAE-PU~\citep{NaVAE} classifier is a classifier based on variational autoencoder designed for PU data. It proved to be one of the most effective recent contributions to modern PU learning due to usage of generated PU examples to offset scarcity of labeled examples for low label frequencies. VAE-PU+OCC~\citep{VAEPUCC} model improves on VAE-PU via more refined choice of artificial PU sample. The modification consists of selecting most likely positive samples from unlabeled dataset instead of using artificially generated examples directly as it is done in VAE-PU~\citep{NaVAE}. For the selection task, usage of one-class-classification (OCC) methods trained on labeled sample was proposed. The modifications significantly improved of the baseline VAE-PU especially in the middle label frequency area, in particular in the case of the two recommended variants -- combining VAE-PU with the $A^3$ and Isolation Forest models, respectively.

    VAE-PU-Bayes aims to further improve upon VAE-PU+OCC performance. Inner selection of the predicted positive examples is a crucial part of the VAE-PU+OCC, but general purpose one-class classifiers are -- on the whole -- a relatively low power methods, as they work with very limited information -- only using the inlier sample distribution. Note that even in the standard PU problem, we can use more information than that, as we have access label information for all of the training examples. This allows for training a classifier which can be used for $s(x)$ estimation. VAE-PU-Bayes combines such a classifier with the VAE-PU $y(x)$ estimation in order to apply $d_{B}^{PU}$ rule. As here the aim is to only filter the unlabeled set, we have $S = 0$ on it and can use the relevant part of the $d_{B}^{PU}$ rule. Note that for the unlabeled stratum, we can rewrite it as follows (see Eq.~(\ref{eq:equiv})):
    \begin{equation*}
        y(x) > \frac{1 + s(x)}{2} \, \equiv \, \frac{y(x) - s(x)}{1 - y(x)} > 1.
    \end{equation*}
    An important consideration is that for numerical reasons, the proportions in the training dataset are crucial to VAE-PU training (refer to~\citep{VAEPUCC} for details) -- due to that, the number of selected likely positives should reflect the true portion on unlabeled positives present in the dataset. Thus, instead of choosing all unlabeled elements satisfying $\frac{y(x) - s(x)}{1 - y(x)} > 1$ as likely positives, we calculate an example's score as $\frac{y(x) - s(x)}{1 - y(x)}$ and select the appropriate number of examples with the highest score as an approximation of internal true PU sample. This approach to PU sample generation is consistent with the decision rule proposed in the paper, and can significantly outperform OCC-based models due to more powerful classification approach.

\section{Numerical experiments}
\label{Numerical}

    To check the effectiveness of the proposed approach, we prepared an extensive suite of experiments. We considered 4 synthetic and 6 real-world datasets:
    \begin{itemize}
        \item All synthetic datasets are generated using a mixture of two 20D Gaussian distributions (with different means $0$ and $\mu$ and unit covariance $I$, except Variant 3) as a feature vector. This implies that indicator $Y$ of an element of a mixture is drawn from the logistic distribution with parameter $\beta$ ($\beta$ is equal to direction of LDA boundary between feature clusters; we use intercept value which ensures $\pi =0.5$). The following variants were used:
        \begin{itemize}
            \item \textbf{Variant 1.} Propensity score for each positive example $e_1(x)$ equals $\sigma(\gamma^Tx+r)$, $\sigma(\cdot)$ being the logistic function, parameter vector $\gamma = [\gamma_1, \gamma_2, ..., \gamma_p] = [0.5, 0.5, ..., 0.5]$) and intercept $r$ is tuned to ensure correct label frequency. Intercept tuning uses the assumed label frequency error as the objective, which is minimized using differential evolution algorithm.
             This allows us to construct synthetic datasets with both required labeling probabilities and label frequencies.
            \item \textbf{Variant 2.} Propensity score: $e_2(x) = e_1(x)^{10}$ which approximates step-wise function and has been considered in \citep{LBE}.
            \item \textbf{Variant 3.} In this variant, covariance matrix is diagonal, non-unit matrix in order to obtain non-logistic data (the diagonal vector equals: $[1, 2, 1, 2, ..., 1, 2]$), $e_3(x) = e_1(x)$,
            \item \textbf{Variant 4 (SCAR).} Constant propensity score, equal to label frequency: $e_4(x) = c$ (equivalent to the SCAR assumption).
        \end{itemize}
        \item Real-world (characteristics of the data sets are given in the Appendix~\ref{appendix:datasets}, their prior probabilities $\pi$ range from 0.4 to 0.53):
        \begin{itemize}
            \item \textbf{MNIST}\footnote{\url{http://yann.lecun.com/exdb/mnist/}} -- two different tasks, 3 versus 5 (images of digit \textit{3} are positive, \textit{5} -- negative, abbreviated to 3v5) and OvE (images of \textit{odd} digits are positive, \textit{even} -- negative),
            \item \textbf{CIFAR-10}\footnote{\url{https://www.cs.toronto.edu/~kriz/cifar.html}} -- two different tasks, CT (\textit{automobile} (car) images are positive, \textit{truck} -- negative) and VA (vehicles (\textit{airplane}, \textit{automobile}, \textit{ship} and \textit{truck}) images are positive; animals (\textit{bird}, \textit{cat}, \textit{deer}, \textit{dog}, \textit{frog} and \textit{horse}) -- negative),
            \item \textbf{STL-10}\footnote{\url{https://cs.stanford.edu/~acoates/stl10/}} -- identical classes (but more complex images) as in CIFAR-10, only VA (Vehicle-Animal) split is considered,
            \item \textbf{CDC-Diabetes}\footnote{\url{https://archive.ics.uci.edu/dataset/891/cdc+diabetes+health+indicators}} -- original class split (rebalanced).
        \end{itemize}
    \end{itemize}

    We performed experiments for multiple label frequencies ($c \in \{0.02, 0.1, 0.3, 0.5, 0.7, 0.9\}$) in order to account for various PU task difficulties and labeling scenarios. To obtain such datasets, we synthetically generated label vectors $S$ corresponding to each label frequency. For synthetic datasets, we use a labeling described above; for real-world datasets, we used feature-based labeling based on examples' properties. For MNIST datasets, examples are labeled depending on digit "boldness" -- a portion of most bold (measured by average pixel value) examples are labeled; for CIFAR-10, a "redness" measure is used -- the most red (according to the measure $r(x) = (R(x) - G(x)) + (R(x) - B(x))$, where $R(\cdot), G(\cdot), B(\cdot)$ correspond to mean R, G and B channel pixel values of input image $x$) examples are labeled; STL-10 uses labeling identical to CIFAR-10. "Maximal value" labeling (taking a portion of the dataset with the highest measure values) as opposed to probabilistic sampling (with probabilities based on those measures) aims to obtain a maximally difficult problem -- note that in that case, labeled positive examples are maximally different from the unlabeled positive examples according to the labeling metric. While this deviates from the probabilistic, propensity score based labeling assumed by the methods, it also helps to measure robustness of the method against assumption violations. CDC-Diabetes aims to simulate a more practical, real-world PU scenario -- there, labeling (diagnosis) probability scales with age (quadratically) and education level (linearly with subsequent stages of education) to model health awareness increase for senior citizens and more educated people.

    We propose the following variants of the three popular no-SCAR PU methods:
    \begin{itemize}
        \item \textbf{LBE+S}. LBE~\citep{LBE} method is a natural candidate due to explicit modeling of both posterior probability $y(x)$ of $Y=1$ and propensity score $e(x)$ (recall that we can obtain posterior probability of $S=1$ by using $s(x) = e(x) y(x)$). After training the LBE classifier, we use both fitted components as plug-in 
        estimators of $y(x)$ and $s(x)$ values in $d_{B}^{PU}$~rule.
        \item \textbf{VAE-PU+S} (abbrev. \textbf{VP+S}). We use VAE-PU~\citep{NaVAE} classifier (described in section~\ref{section:vae-pu-bayes}) as the base. As this model does not natively use the notion of propensity score in contrast to LBE, we introduce a separate feed-forward neural network for $s(x)$ estimation, trained separately from VAE-PU in the additional training step. Its predictions are then fed (together with VAE-PU's $y(x)$ estimations) to the proposed decision rule.
        \item \textbf{VAE-PU-Bayes+S} (abbrev. \textbf{VP-B+S}). We use a newly introduced VAE-PU-Bayes classifier (described in section~\ref{section:vae-pu-bayes}) as the base. Similarly to VAE-PU, $s(x)$ estimator is trained and provided externally using available $(X_i,S_i)_{i=1}^n$ sample.
    \end{itemize}
    Note that for \textrm{synthetic} datasets, we can obtain accurate values of both $y(x)$ and $s(x)$; for those datasets we will additionally show results of the following two pseudo-methods:
    \begin{itemize}
        \item \textbf{S-Prophet}. Corresponds to the application of $d_{B}^{PU}$~rule (\ref{eq:bayes_PU}) with exact $y(x)$ and $s(x)$.
        \item \textbf{Y-Prophet}. Corresponds to a "naive" approach, where a researcher infers $Y = 1$ for test labeled examples with $S = 1$; but then (as one would in the standard PU task) blindly applies ($\ref{eq:bayes_x}$) to all other examples. Note that we assume knowledge of $y(x)$.
    \end{itemize}
    We also define a "naive" versions of LBE+S, VAE-PU+S and VAE-PU-Bayes+S in a similar way (as LBE, VAE-PU and VAE-PU-Bayes) -- by assuming $Y = 1$ for labeled test examples, and using the simple $d_{B}$~rule for the unlabeled examples.

    In order to evaluate the performance, we focus on the "U-metrics", that is metrics calculated for unlabeled stratum. As prediction for labeled test examples is trivial, omitting them in the evaluation results paints clearer picture of the true, underlying decision performance. As an example, U-Accuracy is an Accuracy calculated only on the $S = 0$ stratum: $\textit{U-ACC} = {n_U} ^{-1}\sum_{x_U \in U} \I\{d(x_U, s) = y_U\}$.
    
    We prove the effectiveness of the proposed modification in two steps. First, we show which of the proposed variants (relying on $d_{B}^{PU}$~rule) performs the best on our benchmark tasks. We then go on to compare the best variant with its naive counterpart, showing the benefits of applying our proposed decision rule. Each experiment (defined as a combination of dataset, label frequency and method) was performed 10 times, each time initialized with a different random seed (equal to experiment number). All code used for method implementation and performed experiments is publicly available\footnote{\url{https://github.com/wawrzenczyka/VP-Bayes-S}}. 

    The result section will also contain a brief comparison of VAE-PU-Bayes (abbrev. VP-B) method with the baseline VAE-PU (abbrev. VP) and two recommended VAE-PU+OCC variants -- $A^3$ (abbrev. {VP-$\bm{A^3}$}) and Isolation Forest (abbrev. {VP-IF}). Those experiments were performed without test label availability, and use accuracy as the main metric. The other experimental settings do not differ from the augmented PU prediction experiments. The code for this method is a modification of the original VAE-PU+OCC code, also publicly available in a separate repository\footnote{\url{https://github.com/wawrzenczyka/VAE-PU-Bayes}}.

    \subsection{Results of experiments}

        \begin{table}[tbp]
	\centering \scriptsize \renewcommand{\arraystretch}{1.2}
	\caption{Accuracy values -- VAE-PU-Bayes (traditional PU setting)}
	\label{tab:Accuracy}
\begin{adjustbox}{center}
\scalebox{0.71}{
	\begin{tabular}{ll|llllll}
		\toprule
		\textbf{c} & \textbf{Method} & \textbf{MNIST 3v5} & \textbf{MNIST OvE} & \textbf{CIFAR CT} & \textbf{CIFAR VA} & \textbf{STL VA} & \textbf{CDC Diabetes} \\
		\midrule
		\multirow{4}{*}{0.02} & VP &  \textbf{79.67 $\pm$ 0.90} &           70.00 $\pm$ 1.76 &           87.31 $\pm$ 0.58 &           90.51 $\pm$ 0.52 &           81.64 $\pm$ 0.44 &           50.82 $\pm$ 0.22 \\
		     & VP-$A^3$ &           79.01 $\pm$ 0.70 &           74.89 $\pm$ 1.62 &           83.67 $\pm$ 1.05 &           90.73 $\pm$ 0.27 &           79.62 $\pm$ 0.55 &           53.77 $\pm$ 1.33 \\
		     & VP-IF &           79.07 $\pm$ 0.75 &  \textbf{76.87 $\pm$ 0.99} &           89.98 $\pm$ 1.23 &           89.99 $\pm$ 0.36 &           79.98 $\pm$ 1.06 &           52.38 $\pm$ 1.20 \\
			 \cline{2-8}
		     & VP-B &           78.65 $\pm$ 0.87 &           73.13 $\pm$ 1.70 &  \textbf{91.74 $\pm$ 0.90} &  \textbf{93.70 $\pm$ 0.17} &  \textbf{84.68 $\pm$ 0.65} &  \textbf{57.92 $\pm$ 1.31} \\
		\cline{1-8}
		\multirow{4}{*}{0.10} & VP &           83.57 $\pm$ 0.59 &           77.08 $\pm$ 0.92 &           91.22 $\pm$ 0.19 &           91.54 $\pm$ 0.31 &           85.31 $\pm$ 0.32 &           51.37 $\pm$ 0.25 \\
		     & VP-$A^3$ &           89.91 $\pm$ 0.32 &           83.14 $\pm$ 1.41 &           90.35 $\pm$ 0.47 &           92.35 $\pm$ 0.28 &           84.91 $\pm$ 0.44 &           59.78 $\pm$ 1.36 \\
		     & VP-IF &           90.12 $\pm$ 0.30 &           83.60 $\pm$ 1.28 &           92.26 $\pm$ 0.39 &           90.21 $\pm$ 0.57 &           85.52 $\pm$ 0.56 &           57.24 $\pm$ 1.51 \\
			 \cline{2-8}
		     & VP-B &  \textbf{90.76 $\pm$ 0.54} &  \textbf{85.72 $\pm$ 1.05} &  \textbf{93.73 $\pm$ 0.17} &  \textbf{94.39 $\pm$ 0.14} &  \textbf{88.44 $\pm$ 0.28} &  \textbf{63.46 $\pm$ 0.83} \\
		\cline{1-8}
		\multirow{4}{*}{0.30} & VP &           86.77 $\pm$ 0.48 &           83.71 $\pm$ 0.27 &           92.96 $\pm$ 0.28 &           93.72 $\pm$ 0.13 &           88.23 $\pm$ 0.27 &           54.32 $\pm$ 0.26 \\
		     & VP-$A^3$ &           92.65 $\pm$ 0.22 &           90.49 $\pm$ 0.23 &           89.88 $\pm$ 0.68 &           93.45 $\pm$ 0.12 &           86.38 $\pm$ 0.37 &           68.32 $\pm$ 0.38 \\
		     & VP-IF &           92.73 $\pm$ 0.22 &           90.59 $\pm$ 0.23 &           93.37 $\pm$ 0.21 &           92.02 $\pm$ 0.44 &           87.11 $\pm$ 0.51 &           67.72 $\pm$ 0.44 \\
			 \cline{2-8}
		     & VP-B &  \textbf{92.98 $\pm$ 0.34} &  \textbf{90.88 $\pm$ 0.27} &  \textbf{94.22 $\pm$ 0.13} &  \textbf{94.95 $\pm$ 0.06} &  \textbf{89.99 $\pm$ 0.26} &  \textbf{69.87 $\pm$ 0.19} \\
		\cline{1-8}
		\multirow{4}{*}{0.50} & VP &           88.32 $\pm$ 0.55 &           80.87 $\pm$ 1.35 &           92.91 $\pm$ 0.31 &           88.19 $\pm$ 0.37 &           88.57 $\pm$ 0.49 &           60.58 $\pm$ 0.37 \\
		     & VP-$A^3$ &           93.28 $\pm$ 0.59 &           91.88 $\pm$ 0.42 &           88.03 $\pm$ 1.16 &           93.44 $\pm$ 0.12 &           87.46 $\pm$ 0.24 &           70.97 $\pm$ 0.19 \\
		     & VP-IF &           93.40 $\pm$ 0.55 &           91.59 $\pm$ 0.35 &           93.74 $\pm$ 0.13 &           92.04 $\pm$ 0.26 &           87.91 $\pm$ 0.35 &           70.66 $\pm$ 0.22 \\
			 \cline{2-8}
		     & VP-B &  \textbf{93.92 $\pm$ 0.38} &  \textbf{92.10 $\pm$ 0.28} &  \textbf{94.46 $\pm$ 0.17} &  \textbf{94.99 $\pm$ 0.05} &  \textbf{90.57 $\pm$ 0.30} &  \textbf{71.79 $\pm$ 0.12} \\
		\cline{1-8}
		\multirow{4}{*}{0.70} & VP &           91.58 $\pm$ 0.60 &           91.17 $\pm$ 0.29 &           94.20 $\pm$ 0.20 &           94.67 $\pm$ 0.08 &           90.12 $\pm$ 0.34 &           65.91 $\pm$ 0.25 \\
		     & VP-$A^3$ &           93.89 $\pm$ 0.46 &           94.10 $\pm$ 0.28 &           88.93 $\pm$ 1.41 &           93.99 $\pm$ 0.07 &           89.06 $\pm$ 0.28 &           72.01 $\pm$ 0.07 \\
		     & VP-IF &           94.21 $\pm$ 0.39 &           94.39 $\pm$ 0.25 &           93.99 $\pm$ 0.16 &           93.74 $\pm$ 0.10 &           89.27 $\pm$ 0.32 &           71.93 $\pm$ 0.15 \\
			 \cline{2-8}
		     & VP-B &  \textbf{94.59 $\pm$ 0.57} &  \textbf{94.78 $\pm$ 0.16} &  \textbf{94.51 $\pm$ 0.18} &  \textbf{95.28 $\pm$ 0.04} &  \textbf{91.01 $\pm$ 0.24} &  \textbf{72.42 $\pm$ 0.07} \\
		\cline{1-8}
		\multirow{4}{*}{0.90} & VP &           94.63 $\pm$ 0.17 &           93.15 $\pm$ 0.25 &  \textbf{94.49 $\pm$ 0.14} &           94.79 $\pm$ 0.13 &           91.14 $\pm$ 0.23 &           71.02 $\pm$ 0.17 \\
		     & VP-$A^3$ &           95.35 $\pm$ 0.15 &           95.90 $\pm$ 0.10 &           91.12 $\pm$ 0.34 &           94.69 $\pm$ 0.09 &           91.03 $\pm$ 0.24 &           72.21 $\pm$ 0.07 \\
		     & VP-IF &  \textbf{95.70 $\pm$ 0.18} &           95.80 $\pm$ 0.12 &           94.48 $\pm$ 0.19 &           94.58 $\pm$ 0.08 &  \textbf{91.29 $\pm$ 0.29} &  \textbf{72.45 $\pm$ 0.13} \\
			 \cline{2-8}
		     & VP-B &           95.29 $\pm$ 0.16 &  \textbf{95.96 $\pm$ 0.11} &           94.29 $\pm$ 0.22 &  \textbf{95.12 $\pm$ 0.13} &           91.16 $\pm$ 0.27 &           72.20 $\pm$ 0.13 \\
		\bottomrule
	\end{tabular}
}
\end{adjustbox}
\end{table}

        \noindent \textbf{VAE-PU-Bayes.} First, we show the effectiveness of VAE-PU-Bayes in traditional PU setting. Table~\ref{tab:Accuracy} presents the accuracy comparison between the newly introduced variant and previously existing VAE-PU and VAE-PU+OCC. In the vast majority of cases it outperforms the other VAE-PU variants, often by a very large margin -- up to 5 percentage points (pp.). The only exceptions are the lowest label frequency $c=0.02$, where is it outperformed on MNIST datasets, and $c = 0.9$, where even in this case it is roughly comparable to the best alternative. As VAE-PU+OCC was shown to achieve state-of-the-art level performance when compared to non-generative alternatives \citep{VAEPUCC}, VAE-PU-Bayes can be recommended as an improved variant of this model for traditional PU learning problems.
        
        \vspace{.3cm}
    
        \noindent \textbf{Augmented PU prediction.} The rest of the result section focuses on augmented PU prediction scenario (with available test label). We stress that the aim here is to choose the best performing method among possible proposals for the new scenario.
        Tables~\ref{tab:U-Accuracy-Synth+S} and~\ref{tab:U-Accuracy-Real+S} aggregate experiments performed with $d_{B}^{PU}$~rule for synthetic and real-world datasets, respectively. The best U-Accuracy is marked in bold for each dataset and label frequency combination. The results for Balanced Accuracy are given in the Appendix~\ref{appendix:balanced_accuracy} (note that the ratio 
        of positives to negatives among unlabeled equals $\pi(1-c)/(1-\pi)$ and may be small for $c=0.7,0.9)$.
        For synthetic datasets, VP-B+S is the top performer in the low frequency region; LBE+S does not work well for low label frequencies, but tends to overtake VP-B+S for $c = 0.7$ -- then it levels off and falls off for $c = 0.9$. Even though VP+S is better that VP-B-S for $c = 0.9$, it is outperformed by it for all other label frequencies. For real-world datasets, VP-B+S shows even better performance, dominating in the vast majority of test cases, except for high label frequencies $c=0.5, 0.7$ in the case of CDC Diabetes (where it is outperformed by LBE-S). Overall, we find that VP-B+S is the empirical variant of rule~({\ref{eq:bayes_PU}}) most suited for general recommendation and use, thus we will use it in the further results' presentation. We also note that the dependence of performance on labeling frequency $c$ is much less pronounced here than in classical PU inference. This is due to the fact that large value of $c$ means in general relatively smaller number of positive observations among unlabeled ones and they are harder to detect.

        \begin{table}[tb]
	\centering \scriptsize \renewcommand{\arraystretch}{1.2}
	\caption{U-Accuracy values -- Method comparison -- Synthetic datasets}
	\label{tab:U-Accuracy-Synth+S}
\begin{adjustbox}{center}
\scalebox{0.96}{
	\begin{tabular}{ll|llll}
		\toprule
		\textbf{c} & \textbf{Method} & \textbf{Synth. 1} & \textbf{Synth. 2} & \textbf{Synth. 3} & \textbf{Synth. SCAR} \\
		\midrule
		\multirow{4}{*}{0.02} & S-Prophet &           73.29 $\pm$ 0.35 &           73.24 $\pm$ 0.35 &           71.37 $\pm$ 0.35 &           73.48 $\pm$ 0.35 \\
			& VP+S &           60.55 $\pm$ 2.48 &           59.15 $\pm$ 2.62 &           59.77 $\pm$ 2.40 &           63.19 $\pm$ 1.75 \\
			& VP-B+S &  \textbf{61.23 $\pm$ 2.35} &  \textbf{59.35 $\pm$ 2.66} &  \textbf{60.16 $\pm$ 2.36} &  \textbf{63.45 $\pm$ 1.82} \\
			& LBE+S &           50.32 $\pm$ 0.50 &           50.59 $\pm$ 0.50 &           50.66 $\pm$ 0.49 &           50.29 $\pm$ 0.47 \\
		\cline{1-6}
		\multirow{4}{*}{0.10} & S-Prophet &           72.63 $\pm$ 0.30 &           72.16 $\pm$ 0.35 &           70.61 $\pm$ 0.30 &           73.74 $\pm$ 0.34 \\
			& VP+S &           67.18 $\pm$ 0.42 &           65.96 $\pm$ 0.58 &           67.02 $\pm$ 0.57 &           67.64 $\pm$ 0.42 \\
			& VP-B+S &  \textbf{67.71 $\pm$ 0.49} &  \textbf{66.63 $\pm$ 0.60} &  \textbf{67.49 $\pm$ 0.59} &  \textbf{68.37 $\pm$ 0.50} \\
			& LBE+S &           52.72 $\pm$ 0.47 &           53.45 $\pm$ 0.50 &           53.04 $\pm$ 0.45 &           52.39 $\pm$ 0.53 \\
		\cline{1-6}
		\multirow{4}{*}{0.30} & S-Prophet &           71.70 $\pm$ 0.42 &           70.83 $\pm$ 0.48 &           69.45 $\pm$ 0.39 &           74.30 $\pm$ 0.46 \\
			& VP+S &           67.77 $\pm$ 0.57 &           65.29 $\pm$ 0.64 &           66.90 $\pm$ 0.55 &           70.20 $\pm$ 0.45 \\
			& VP-B+S &  \textbf{68.51 $\pm$ 0.54} &  \textbf{66.41 $\pm$ 0.57} &  \textbf{67.27 $\pm$ 0.47} &  \textbf{71.03 $\pm$ 0.42} \\
			& LBE+S &           61.05 $\pm$ 0.36 &           60.80 $\pm$ 0.43 &           61.03 $\pm$ 0.31 &           58.80 $\pm$ 0.52 \\
		\cline{1-6}
		\multirow{4}{*}{0.50} & S-Prophet &           72.78 $\pm$ 0.57 &           71.96 $\pm$ 0.46 &           70.75 $\pm$ 0.59 &           76.93 $\pm$ 0.56 \\
			& VP+S &           66.87 $\pm$ 0.41 &           65.04 $\pm$ 0.47 &           66.07 $\pm$ 0.67 &           69.78 $\pm$ 0.68 \\
			& VP-B+S &           67.90 $\pm$ 0.45 &           65.57 $\pm$ 0.46 &           67.01 $\pm$ 0.51 &  \textbf{72.31 $\pm$ 0.38} \\
			& LBE+S &  \textbf{68.72 $\pm$ 0.51} &  \textbf{67.72 $\pm$ 0.50} &  \textbf{68.45 $\pm$ 0.45} &           70.86 $\pm$ 0.48 \\
		\cline{1-6}
		\multirow{4}{*}{0.70} & S-Prophet &           78.79 $\pm$ 0.40 &           78.37 $\pm$ 0.35 &           77.70 $\pm$ 0.50 &           81.31 $\pm$ 0.37 \\
			& VP+S &           67.28 $\pm$ 0.88 &           66.38 $\pm$ 0.88 &           66.05 $\pm$ 0.78 &           69.34 $\pm$ 1.27 \\
			& VP-B+S &           70.49 $\pm$ 0.54 &           68.91 $\pm$ 0.51 &           69.04 $\pm$ 0.48 &           73.57 $\pm$ 0.59 \\
			& LBE+S &  \textbf{74.74 $\pm$ 0.42} &  \textbf{73.50 $\pm$ 0.52} &  \textbf{73.57 $\pm$ 0.42} &  \textbf{81.03 $\pm$ 0.38} \\
		\cline{1-6}
		\multirow{4}{*}{0.90} & S-Prophet &           91.20 $\pm$ 0.49 &           91.26 $\pm$ 0.50 &           91.42 $\pm$ 0.44 &           91.83 $\pm$ 0.36 \\
			& VP+S &  \textbf{85.54 $\pm$ 0.49} &  \textbf{86.00 $\pm$ 0.76} &  \textbf{85.64 $\pm$ 0.70} &  \textbf{87.97 $\pm$ 0.55} \\
			& VP-B+S &           84.00 $\pm$ 0.42 &           84.48 $\pm$ 0.69 &           83.90 $\pm$ 0.56 &           86.50 $\pm$ 0.54 \\
			& LBE+S &           74.76 $\pm$ 0.46 &           74.53 $\pm$ 0.47 &           73.57 $\pm$ 0.55 &           78.14 $\pm$ 0.48 \\
		\bottomrule
	\end{tabular}
}
\end{adjustbox}
\end{table}

        \begin{table}[tb]
	\centering \scriptsize \renewcommand{\arraystretch}{1.2}
	\caption{U-Accuracy values -- Method comparison -- Real-world datasets}
	\label{tab:U-Accuracy-Real+S}
\begin{adjustbox}{center}
\scalebox{0.70}{
	\begin{tabular}{ll|llllll}
		\toprule
		\textbf{c} & \textbf{Method} & \textbf{MNIST 3v5} & \textbf{MNIST OvE} & \textbf{CIFAR CT} & \textbf{CIFAR VA} & \textbf{STL VA} & \textbf{CDC-Diabetes} \\
		\midrule
		\multirow{3}{*}{0.02} & VP+S &           77.74 $\pm$ 1.10 &           68.47 $\pm$ 1.18 &           87.19 $\pm$ 0.49 &           90.32 $\pm$ 0.25 &           81.43 $\pm$ 0.66 &           49.76 $\pm$ 1.49 \\
		     & VP-B+S &  \textbf{78.74 $\pm$ 1.53} &  \textbf{74.91 $\pm$ 1.51} &  \textbf{92.45 $\pm$ 0.43} &  \textbf{94.11 $\pm$ 0.09} &  \textbf{84.54 $\pm$ 0.67} &  \textbf{51.15 $\pm$ 1.74} \\
		     & LBE+S &           47.38 $\pm$ 0.32 &           49.82 $\pm$ 0.14 &           50.50 $\pm$ 0.40 &           60.67 $\pm$ 0.22 &           60.55 $\pm$ 0.28 &           50.47 $\pm$ 0.20 \\
		\cline{1-8}
		\multirow{3}{*}{0.10} & VP+S &           80.21 $\pm$ 0.61 &           73.96 $\pm$ 1.43 &           91.42 $\pm$ 0.33 &           91.93 $\pm$ 0.33 &           86.36 $\pm$ 0.38 &           56.57 $\pm$ 0.79 \\
		     & VP-B+S &  \textbf{84.32 $\pm$ 0.76} &  \textbf{83.12 $\pm$ 1.24} &  \textbf{93.45 $\pm$ 0.19} &  \textbf{94.37 $\pm$ 0.12} &  \textbf{88.74 $\pm$ 0.30} &  \textbf{61.01 $\pm$ 0.69} \\
		     & LBE+S &           49.81 $\pm$ 0.34 &           51.55 $\pm$ 0.15 &           53.19 $\pm$ 0.39 &           62.81 $\pm$ 0.29 &           62.93 $\pm$ 0.28 &           52.55 $\pm$ 0.20 \\
		\cline{1-8}
		\multirow{3}{*}{0.30} & VP+S &           80.66 $\pm$ 0.73 &           78.38 $\pm$ 0.96 &           92.95 $\pm$ 0.30 &           93.49 $\pm$ 0.16 &           88.93 $\pm$ 0.20 &           51.76 $\pm$ 0.94 \\
		     & VP-B+S &  \textbf{86.95 $\pm$ 0.52} &  \textbf{87.98 $\pm$ 0.64} &  \textbf{94.32 $\pm$ 0.14} &  \textbf{95.22 $\pm$ 0.07} &  \textbf{89.90 $\pm$ 0.30} &  \textbf{62.63 $\pm$ 0.89} \\
		     & LBE+S &           56.26 $\pm$ 0.34 &           57.11 $\pm$ 0.15 &           63.07 $\pm$ 1.04 &           74.37 $\pm$ 2.36 &           71.53 $\pm$ 1.13 &           58.72 $\pm$ 0.20 \\
		\cline{1-8}
		\multirow{3}{*}{0.50} & VP+S &           82.25 $\pm$ 0.78 &           81.15 $\pm$ 0.93 &           94.39 $\pm$ 0.20 &           94.63 $\pm$ 0.26 &           90.85 $\pm$ 0.26 &           48.01 $\pm$ 0.85 \\
		     & VP-B+S &  \textbf{89.20 $\pm$ 0.81} &  \textbf{90.25 $\pm$ 0.56} &  \textbf{95.13 $\pm$ 0.20} &  \textbf{95.65 $\pm$ 0.12} &  \textbf{91.44 $\pm$ 0.30} &           66.88 $\pm$ 0.37 \\
		     & LBE+S &           64.23 $\pm$ 0.33 &           64.80 $\pm$ 0.23 &           80.81 $\pm$ 1.68 &           85.35 $\pm$ 1.18 &           84.15 $\pm$ 1.00 &  \textbf{72.39 $\pm$ 0.81} \\
		\cline{1-8}
		\multirow{3}{*}{0.70} & VP+S &           86.42 $\pm$ 0.65 &           85.88 $\pm$ 0.67 &           95.32 $\pm$ 0.22 &           95.53 $\pm$ 0.18 &  \textbf{93.27 $\pm$ 0.29} &           42.13 $\pm$ 1.05 \\
		     & VP-B+S &  \textbf{92.19 $\pm$ 0.47} &  \textbf{92.76 $\pm$ 0.43} &  \textbf{95.73 $\pm$ 0.18} &  \textbf{96.23 $\pm$ 0.12} &  \textbf{93.27 $\pm$ 0.21} &           71.74 $\pm$ 0.49 \\
		     & LBE+S &           74.84 $\pm$ 0.40 &           76.65 $\pm$ 0.22 &           93.86 $\pm$ 0.57 &           95.16 $\pm$ 0.25 &           92.17 $\pm$ 0.38 &  \textbf{77.12 $\pm$ 0.79} \\
		\cline{1-8}
		\multirow{3}{*}{0.90} & VP+S &           91.90 $\pm$ 0.35 &           90.91 $\pm$ 0.35 &  \textbf{96.65 $\pm$ 0.16} &           96.76 $\pm$ 0.15 &  \textbf{95.83 $\pm$ 0.19} &           42.56 $\pm$ 2.15 \\
		     & VP-B+S &  \textbf{94.07 $\pm$ 0.25} &  \textbf{95.73 $\pm$ 0.23} &           96.47 $\pm$ 0.16 &  \textbf{97.11 $\pm$ 0.09} &           95.37 $\pm$ 0.24 &  \textbf{83.53 $\pm$ 0.33} \\
		     & LBE+S &           90.00 $\pm$ 0.27 &           87.03 $\pm$ 0.97 &           94.21 $\pm$ 0.48 &           94.69 $\pm$ 1.25 &           94.01 $\pm$ 0.85 &           71.09 $\pm$ 1.63 \\
		\bottomrule
	\end{tabular}
}
\end{adjustbox}
\end{table}

        In Tables~\ref{tab:U-Accuracy-Synth-Compare} and~\ref{tab:U-Accuracy-Real-Compare}, we aim to capture the impact of using $d_{B}^{PU}$~rule-based VP-B+S instead of its naive counterpart, VP-B. For synthetic datasets, where we have access to true $y(x)$ and $s(x)$ value, we contrast them with the analogous, reference S-Prophet and Y-Prophet methods. In this case, we also introduce the "semi-Prophet" methods -- VP-B+S with true $s(x)$ and VP-B+S with true $y(x)$, where the true values replace the corresponding VP-B+S probability estimation. First thing to note is that S-Prophet is nearly equivalent to Y-Prophet in low label frequency setting. When there is a small number of labeled examples, it leads to low in expectation predicted labeling probability $s(x)$. As the rules $d_{B}^{PU}$ and $d_{B}$ are equivalent when $s(x) = 0$, for low label frequencies the change in predicted class is relatively infrequent. As the label frequency increases, so does the discrepancy between prophet methods -- culminating in the drastic difference of 20 pp. for $c = 0.9$. The differences between VP-B+S and VP-B are not as big, and also tend to increase jointly with label frequency. 
        However, p-value of the binomial test for testing $H_0$: P(U-acc. of VP-B $>$ U-acc. of VP-B+S)$\geq 1/2$ against the opposite hypothesis, equals to $1.8\times 10^{-5}$ (corresponding to 2 wins in 24 trials) for Table ~\ref{tab:U-Accuracy-Synth-Compare} and $1.1\times 10^{-7}$ in case of Table ~\ref{tab:U-Accuracy-Real-Compare}. 
        Using the correct decision rule via VP-B-S we obtain U-Accuracy increase in almost every test scenario, though the margin here is much smaller than in the case of Prophets, and more pronounced for synthetic datasets. Inspecting semi-Prophet results gives us additional insights into the $d_{B}^{PU}$ components. Note that when using true $y(x)$, the VP-B+S semi-Prophet's accuracy does not deviate significantly from S-Prophet's -- even though the estimation of $s(x)$ was fairly crude, it is good enough when combined with accurate $y(x)$ estimations to improve results significantly. The same does not hold true for VP-B+S semi-Prophet using true $s(x)$ values, which indicates that $y(x)$ estimation inaccuracy is a major contributor to the performance drop compared with the Prophet methods. This is evident by contrasting the results with the S-Prophet -- Y-Prophet pair, where using true $s(x)$ for $d_{B}^{PU}$ rule proved to increase performance dramatically for high label frequencies. Note that sometimes even a slight variation of $y(x)$ might led to a between $d_{B}^{PU}$ influencing the final example label or leaving it unchanged.

        \begin{table}[tb]
	\centering \scriptsize \renewcommand{\arraystretch}{1.2}
	\caption{U-Accuracy values -- Decision rule comparison -- Synthetic datasets}
	\label{tab:U-Accuracy-Synth-Compare}
\begin{adjustbox}{center}
\scalebox{0.86}{
	\begin{tabular}{ll|llll}
		\toprule
		\textbf{c} & \textbf{Method} & \textbf{Synth. 1} & \textbf{Synth. 2} & \textbf{Synth. 3} & \textbf{Synth. SCAR} \\
		\midrule
		\multirow{6}{*}{0.02} & S-Prophet &           73.29 $\pm$ 0.35 &           73.24 $\pm$ 0.35 &           71.37 $\pm$ 0.35 &           73.48 $\pm$ 0.35 \\
		     & Y-Prophet &           73.31 $\pm$ 0.36 &           73.24 $\pm$ 0.35 &           71.40 $\pm$ 0.36 &           73.50 $\pm$ 0.35 \\
			 \cline{2-6}
		     & VP-B &           61.00 $\pm$ 2.40 &  \textbf{59.62 $\pm$ 2.56} &           60.14 $\pm$ 2.38 &           63.37 $\pm$ 1.77 \\
		     & VP-B+S &  \textbf{61.23 $\pm$ 2.35} &           59.35 $\pm$ 2.66 &  \textbf{60.16 $\pm$ 2.36} &  \textbf{63.45 $\pm$ 1.82} \\
			 \cline{2-6}
		     & VP-B+S + true s(x) &           60.65 $\pm$ 2.41 &           59.15 $\pm$ 2.69 &           59.98 $\pm$ 2.39 &           63.14 $\pm$ 1.76 \\
		     & VP-B+S + true y(x) &           73.29 $\pm$ 0.37 &           73.21 $\pm$ 0.35 &           71.44 $\pm$ 0.35 &           73.46 $\pm$ 0.33 \\
		\cline{1-6}
		\multirow{6}{*}{0.10} & S-Prophet &           72.63 $\pm$ 0.30 &           72.16 $\pm$ 0.35 &           70.61 $\pm$ 0.30 &           73.74 $\pm$ 0.34 \\
		     & Y-Prophet &           72.63 $\pm$ 0.35 &           72.19 $\pm$ 0.37 &           70.68 $\pm$ 0.37 &           73.66 $\pm$ 0.33 \\
			 \cline{2-6}
		     & VP-B &  \textbf{67.81 $\pm$ 0.48} &           66.42 $\pm$ 0.53 &           67.38 $\pm$ 0.60 &           68.35 $\pm$ 0.48 \\
		     & VP-B+S &           67.71 $\pm$ 0.49 &  \textbf{66.63 $\pm$ 0.60} &  \textbf{67.49 $\pm$ 0.59} &  \textbf{68.37 $\pm$ 0.50} \\
			 \cline{2-6}
		     & VP-B+S + true s(x) &           67.64 $\pm$ 0.50 &           66.33 $\pm$ 0.51 &           67.16 $\pm$ 0.62 &           68.07 $\pm$ 0.48 \\
		     & VP-B+S + true y(x) &           72.71 $\pm$ 0.34 &           71.92 $\pm$ 0.39 &           70.61 $\pm$ 0.35 &           73.73 $\pm$ 0.32 \\
		\cline{1-6}
		\multirow{6}{*}{0.30} & S-Prophet &           71.70 $\pm$ 0.42 &           70.83 $\pm$ 0.48 &           69.45 $\pm$ 0.39 &           74.30 $\pm$ 0.46 \\
		     & Y-Prophet &           71.06 $\pm$ 0.39 &           70.08 $\pm$ 0.39 &           69.00 $\pm$ 0.37 &           73.56 $\pm$ 0.34 \\
			 \cline{2-6}
		     & VP-B &           68.25 $\pm$ 0.47 &           66.27 $\pm$ 0.62 &           67.14 $\pm$ 0.52 &           70.56 $\pm$ 0.43 \\
		     & VP-B+S &  \textbf{68.51 $\pm$ 0.54} &  \textbf{66.41 $\pm$ 0.57} &  \textbf{67.27 $\pm$ 0.47} &  \textbf{71.03 $\pm$ 0.42} \\
			 \cline{2-6}
		     & VP-B+S + true s(x) &           68.19 $\pm$ 0.54 &           66.02 $\pm$ 0.63 &           67.06 $\pm$ 0.51 &           70.72 $\pm$ 0.44 \\
		     & VP-B+S + true y(x) &           71.26 $\pm$ 0.46 &           70.56 $\pm$ 0.52 &           69.19 $\pm$ 0.43 &           74.32 $\pm$ 0.48 \\
		\cline{1-6}
		\multirow{6}{*}{0.50} & S-Prophet &           72.78 $\pm$ 0.57 &           71.96 $\pm$ 0.46 &           70.75 $\pm$ 0.59 &           76.93 $\pm$ 0.56 \\
		     & Y-Prophet &           69.87 $\pm$ 0.40 &           68.83 $\pm$ 0.39 &           67.81 $\pm$ 0.43 &           73.26 $\pm$ 0.35 \\
			 \cline{2-6}
		     & VP-B &           67.07 $\pm$ 0.40 &           65.18 $\pm$ 0.45 &           66.14 $\pm$ 0.65 &           70.43 $\pm$ 0.46 \\
		     & VP-B+S &  \textbf{67.90 $\pm$ 0.45} &  \textbf{65.57 $\pm$ 0.46} &  \textbf{67.01 $\pm$ 0.51} &  \textbf{72.31 $\pm$ 0.38} \\
			 \cline{2-6}
		     & VP-B+S + true s(x) &           67.58 $\pm$ 0.38 &           65.36 $\pm$ 0.51 &           66.67 $\pm$ 0.63 &           71.99 $\pm$ 0.46 \\
		     & VP-B+S + true y(x) &           72.06 $\pm$ 0.59 &           71.11 $\pm$ 0.61 &           69.83 $\pm$ 0.62 &           76.34 $\pm$ 0.62 \\
		\cline{1-6}
		\multirow{6}{*}{0.70} & S-Prophet &           78.79 $\pm$ 0.40 &           78.37 $\pm$ 0.35 &           77.70 $\pm$ 0.50 &           81.31 $\pm$ 0.37 \\
		     & Y-Prophet &           69.39 $\pm$ 0.41 &           68.79 $\pm$ 0.43 &           67.44 $\pm$ 0.47 &           73.42 $\pm$ 0.35 \\
			 \cline{2-6}
		     & VP-B &           66.46 $\pm$ 0.59 &           65.69 $\pm$ 0.56 &           65.25 $\pm$ 0.65 &           68.76 $\pm$ 0.71 \\
		     & VP-B+S &  \textbf{70.49 $\pm$ 0.54} &  \textbf{68.91 $\pm$ 0.51} &  \textbf{69.04 $\pm$ 0.48} &  \textbf{73.57 $\pm$ 0.59} \\
			 \cline{2-6}
		     & VP-B+S + true s(x) &           69.95 $\pm$ 0.58 &           69.16 $\pm$ 0.49 &           68.88 $\pm$ 0.53 &           73.09 $\pm$ 0.58 \\
		     & VP-B+S + true y(x) &           77.42 $\pm$ 0.45 &           77.11 $\pm$ 0.39 &           75.95 $\pm$ 0.61 &           80.46 $\pm$ 0.46 \\		\cline{1-6}
		\multirow{6}{*}{0.90} & S-Prophet &           91.20 $\pm$ 0.49 &           91.26 $\pm$ 0.50 &           91.42 $\pm$ 0.44 &           91.83 $\pm$ 0.36 \\
		     & Y-Prophet &           71.30 $\pm$ 0.44 &           71.17 $\pm$ 0.45 &           69.25 $\pm$ 0.47 &           73.33 $\pm$ 0.48 \\
			 \cline{2-6}
		     & VP-B &           69.71 $\pm$ 0.37 &           69.47 $\pm$ 0.41 &           68.16 $\pm$ 0.49 &           71.76 $\pm$ 0.42 \\
		     & VP-B+S &  \textbf{84.00 $\pm$ 0.42} &  \textbf{84.48 $\pm$ 0.69} &  \textbf{83.90 $\pm$ 0.56} &  \textbf{86.50 $\pm$ 0.54} \\
			 \cline{2-6}
		     & VP-B+S + true s(x) &           84.12 $\pm$ 0.56 &           83.89 $\pm$ 0.55 &           83.56 $\pm$ 0.65 &           87.03 $\pm$ 0.49 \\
		     & VP-B+S + true y(x) &           90.44 $\pm$ 0.41 &           90.69 $\pm$ 0.47 &           89.72 $\pm$ 0.37 &           90.82 $\pm$ 0.24 \\
		\bottomrule
	\end{tabular}
}
\end{adjustbox}
\end{table}
    
        \begin{table}[tb]
	\centering \scriptsize \renewcommand{\arraystretch}{1.2}
	\caption{U-Accuracy values -- Decision rule comparison -- Real-world datasets}
	\label{tab:U-Accuracy-Real-Compare}
\begin{adjustbox}{center}
	\scalebox{0.69}{
		\begin{tabular}{ll|llllll}
			\toprule
			\textbf{c} & \textbf{Method} & \textbf{MNIST 3v5} & \textbf{MNIST OvE} & \textbf{CIFAR CT} & \textbf{CIFAR MA} & \textbf{STL MA} & \textbf{CDC-Diabetes} \\
			\midrule
			\multirow{2}{*}{0.02} & VP-B &  \textbf{78.75 $\pm$ 1.44} &           74.53 $\pm$ 1.49 &           92.40 $\pm$ 0.41 &           93.94 $\pm$ 0.10 &           84.50 $\pm$ 0.66 &           51.07 $\pm$ 1.76 \\
				& VP-B+S &           78.74 $\pm$ 1.53 &  \textbf{74.91 $\pm$ 1.51} &  \textbf{92.45 $\pm$ 0.43} &  \textbf{94.11 $\pm$ 0.09} &  \textbf{84.54 $\pm$ 0.67} &  \textbf{51.15 $\pm$ 1.74} \\
			\cline{1-8}
			\multirow{2}{*}{0.10} & VP-B &           84.14 $\pm$ 0.65 &           82.67 $\pm$ 1.30 &           93.32 $\pm$ 0.18 &           94.29 $\pm$ 0.12 &           88.54 $\pm$ 0.30 &  \textbf{61.25 $\pm$ 0.80} \\
				& VP-B+S &  \textbf{84.32 $\pm$ 0.76} &  \textbf{83.12 $\pm$ 1.24} &  \textbf{93.45 $\pm$ 0.19} &  \textbf{94.37 $\pm$ 0.12} &  \textbf{88.74 $\pm$ 0.30} &           61.01 $\pm$ 0.69 \\
			\cline{1-8}
			\multirow{2}{*}{0.30} & VP-B &           86.64 $\pm$ 0.56 &           87.89 $\pm$ 0.65 &           94.18 $\pm$ 0.17 &           95.11 $\pm$ 0.07 &  \textbf{90.11 $\pm$ 0.26} &  \textbf{63.44 $\pm$ 0.81} \\
				& VP-B+S &  \textbf{86.95 $\pm$ 0.52} &  \textbf{87.98 $\pm$ 0.64} &  \textbf{94.32 $\pm$ 0.14} &  \textbf{95.22 $\pm$ 0.07} &           89.90 $\pm$ 0.30 &           62.63 $\pm$ 0.89 \\
			\cline{1-8}
			\multirow{2}{*}{0.50} & VP-B &           88.75 $\pm$ 0.84 &           90.19 $\pm$ 0.54 &           94.87 $\pm$ 0.22 &           95.44 $\pm$ 0.11 &           91.35 $\pm$ 0.25 &           66.52 $\pm$ 0.31 \\
				& VP-B+S &  \textbf{89.20 $\pm$ 0.81} &  \textbf{90.25 $\pm$ 0.56} &  \textbf{95.13 $\pm$ 0.20} &  \textbf{95.65 $\pm$ 0.12} &  \textbf{91.44 $\pm$ 0.30} &  \textbf{66.88 $\pm$ 0.37} \\
			\cline{1-8}
			\multirow{2}{*}{0.70} & VP-B &           91.84 $\pm$ 0.48 &           92.73 $\pm$ 0.39 &           95.30 $\pm$ 0.20 &           95.94 $\pm$ 0.10 &           92.64 $\pm$ 0.24 &           67.73 $\pm$ 0.43 \\
				& VP-B+S &  \textbf{92.19 $\pm$ 0.47} &  \textbf{92.76 $\pm$ 0.43} &  \textbf{95.73 $\pm$ 0.18} &  \textbf{96.23 $\pm$ 0.12} &  \textbf{93.27 $\pm$ 0.21} &  \textbf{71.74 $\pm$ 0.49} \\
			\cline{1-8}
			\multirow{2}{*}{0.90} & VP-B &           93.90 $\pm$ 0.25 &           95.45 $\pm$ 0.21 &           95.68 $\pm$ 0.16 &           96.51 $\pm$ 0.05 &           93.54 $\pm$ 0.26 &           68.13 $\pm$ 0.34 \\
				& VP-B+S &  \textbf{94.07 $\pm$ 0.25} &  \textbf{95.73 $\pm$ 0.23} &  \textbf{96.47 $\pm$ 0.16} &  \textbf{97.11 $\pm$ 0.09} &  \textbf{95.37 $\pm$ 0.24} &  \textbf{83.53 $\pm$ 0.33} \\
			\bottomrule
		\end{tabular}
	}
\end{adjustbox}
\end{table}

\begin{figure}[htb]
            \centering
            \includegraphics[width=.475\textwidth]{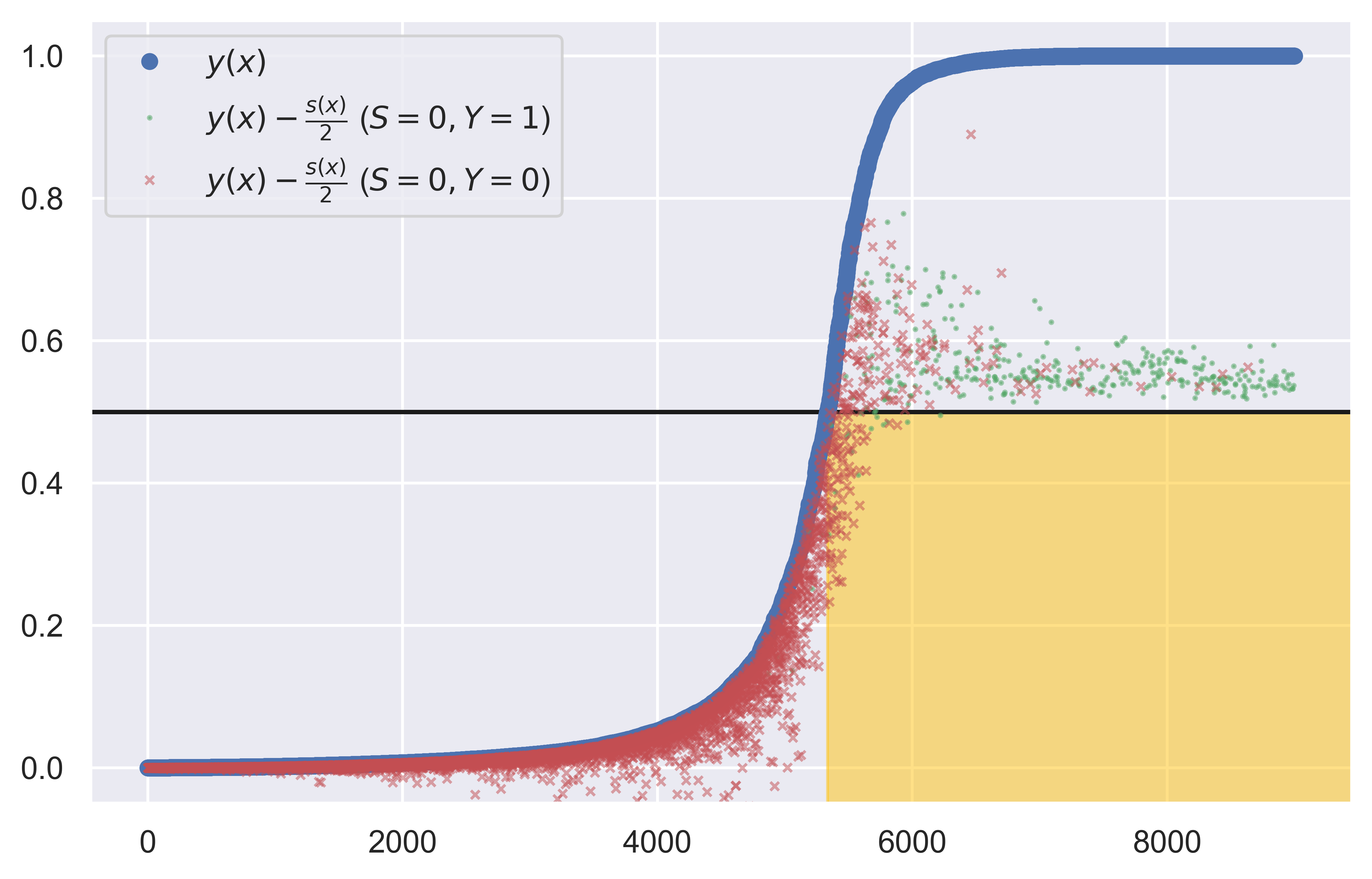}
            \caption{Classification rules for test instances, CIFAR VA, $c = 0.9$, $S = 0$ stratum (colored by test class).}
            \label{fig:problems:s-split}
        \end{figure}

        Results above shows that in real-world scenarios, the performance gain obtained by using the proposed decision rule over $d_B$ rule is systematic but relatively small; this is especially apparent when comparing it to Prophets' improvements. Figure~\ref{fig:problems:s-split} aims to illustrate one of the potential causes of that problem. For the sake of this example, we will plot those values only for samples from $S = 0$ stratum. Note that for this stratum, $d_B$ rule is equivalent to ${\I\{y(x) > 0.5\}}$, whereas $d_B^{PU}$ -- to ${\I\{y(x) - \frac{s(x)}{2} > 0.5\}}$. This formulation provides us with a matching threshold $0.5$ for both rules.

        The example orders the test samples according to increasing $y(x)$ (blue color, basis of $d_B$ rule). In the figure, we introduce one additional dot for each test instance, which now corresponds to $y(x) - \frac{s(x)}{2}$ (basis of $d_B^{PU}$ rule). Those dots are colored based on their test class (positive examples in green, and negative -- in red). As $y(x) - \frac{s(x)}{2}$ is always lower or equal to $y(x)$, in order for the $d_B$ and $d_B^{PU}$ classification rules to differ on the $S = 0$ stratum (i) the blue dot for the given test example must be above black boundary line ($y = 0.5$), and (ii) the other dot (green or red) must be lying below it. The area in the chart where this is possible is shaded gold, and only examples falling there fulfill both conditions. The important thing to note is that the amount of examples falling in the golden area is relatively small, due to approximated $y(x)$ tending to the extremes of $0$ and $1$~--~which might not hold true for the true $y(x)$ distribution. This limits the benefits of applying the $d_B^{PU}$ rule, as even though the examples in the golden area are mostly negative (resulting in decreasing of the number of false signals), and the green, positive unlabeled samples are concentrated in the high $y(x)$ area, the limited number of affected samples by rule's modification lowers the impact of the correction on the metrics such as U-Accuracy.

\section{Conclusions}

    The contribution of this paper is twofold: firstly, we highlight a previously unexplored area of PU learning (augmented PU prediction) where samples' labels at prediction time are available. Secondly, we propose a novel $d_B^{PU}$ decision rule tailored for this setting. We study the basic properties of the proposed rule and contrast it with the properties of the usual Bayes rule based solely on samples' features. We also show that $d_B^{PU}$'s usefulness is not limited to the augmented PU prediction scenario, and it can be employed also in e.g. in traditional PU setting as a part of VAE-PU-Bayes model. The latter half of the paper focuses on the practical experiments, combining $d_B^{PU}$ rule with preexisting PU models. We start off by showing the substantial improvements of VP-B over the VAE-PU+OCC baseline for traditional PU tasks. In augmented PU prediction setting, we identify VP-B+S model as the most promising among the newly constructed methods. By comparing it with its naive counterpart, VP-B, we show that using $d_B^{PU}$ rule systematically improves accuracy on the test dataset. However, results for the two Prophet methods (which use perfect knowledge of $y(x)$ and $s(x)$), as well as semi-Prophets (utilizing the perfect knowledge of only one of those variables) indicate that those improvements could be potentially be significantly larger, especially for the high label frequencies. As this paper introduces a new, practical setting for PU data, it naturally presents researchers with a rich opportunities for further work. One of those possibilities involves better modeling of $y(x)$, which currently is not sensitive enough to corrections via $d_B^{PU}$ rule in direct, practical applications. Proposing new classifiers relying on estimators of both $y(x)$ and $s(x)$ directly (similarly to LBE) is an important challenge. Moreover, an excellent performance of VAE-PU-Bayes in traditional setting encourages further work on this model, or incorporating $d_B^{PU}$ rule as a component of more PU models.



\clearpage
\bibliography{References}
\clearpage


\begin{appendices}
    \setcounter{table}{0}
    \renewcommand{\thetable}{A\arabic{table}}
    
    \section{Example 1 -- full derivation}
    \label{appendix:example_1}
    
        \begin{example}
            Let $y(x) = \Phi(x), X \sim N(0, 1)$, and $x \in \R$ (univariate probit model with standard normal predictor), and let propensity score $e_a(x) = \I \{x > a\}$ i.e. above threshold $a \in \R$ all positive observations are labeled. It is easy to check that
            \begin{equation*}
                \begin{aligned}
                    P(Y = 1)
                        & = \int P(Y = 1|X = x) f(x) \, \dif x 
                        = \int_{-\infty}^\infty \Phi(x) \phi(x) \, \dif x \\
                        & = \int_0^1 z \, \dif z
                        = \frac{1}{2}.                
                \end{aligned}
            \end{equation*}
            and Bayes risk of $d_B(x)$ equals
            \begin{equation*}
                \begin{aligned}
                    L^* 
                        & = \int_{-\infty}^\infty \min{\big( \Phi(x), 1 - \Phi(x) \big)} \phi(x) \, \dif x \\
                        & = \int_{-\infty}^0 \Phi(x) \phi(x) \, \dif x 
                            + \int_0^\infty (1 - \Phi(x)) \phi(x) \, \dif x 
                            = \frac{1}{4}.
                \end{aligned}
            \end{equation*}
            As $s(x) = y(x) \I\{x > a\}$, probability of labeling equals 
            \begin{equation*}
                \begin{aligned}
                    P(S = 1) 
                        & = \int_{-\infty}^\infty P(S = 1|X = x)f(x) \, \dif x
                        = \int_{-\infty}^\infty s(x) \phi(x) \, \dif x \\
                        & = \int_a^\infty \Phi(x) \phi(x) \, \dif x
                        = \frac{1}{2} (1 - \Phi^2(a))
                \end{aligned}
            \end{equation*}
            and $P(S = 0) = \frac{1}{2} (1 + \Phi^2(a))$. Moreover, $L_{PU}^{*} = L_{PU}^{*}(a)$ equals for $a > 0$
            \begin{equation*}
                \begin{aligned}
                    L_{PU}^{*}(a)
                        & = \E_{X, S = 0} \min{\big( \tilde{y}(X, 0), 1 - \tilde{y}(X, 0) \big)} \\
                        & = \left\{
                            \int_{-\infty}^0 \Phi(x) \phi(x) \, \dif x 
                            + \int_0^a (1 - \Phi(x)) \phi(x) \, \dif x 
                        \right\} \\
                        & = \Phi(a) - \frac{\Phi^2(a)}{2} - \frac{1}{4}.
                \end{aligned}
            \end{equation*}
            and analogous calculation for $a\leq 0$ yields $ L_{PU}^{*}(a)=\Phi^2(a)/2$.
            
            Thus the excess risk of $d_B(x)$ defined in (\ref{eq:bayes_x}) for $a > 0$ equals
            \begin{equation*}
                \begin{aligned}
                    & \E_X \big[\min{\big( y(X), 1 - y(X) \big)} \big] \\
                        & \phantom{= \ } - \E_{X, S} \big[\min{\big( \tilde{y}(X, S), 1 - \tilde{y}(X, S) \big)} \big] \\
                    & = \frac{1}{2} - \Phi(a) + \frac{\Phi^2(a)}{2}
                        = \frac{1}{2} \big(\Phi(a) - 1 \big)^2 \geq 0,
                \end{aligned}
            \end{equation*}
            and for $a<0$ equals $\frac{1}{4} -\frac{\Phi^2(a)}{2}\geq 0$. Note that for $a\to \infty$ excess risk tends to 0 as $P_{X, S = 0}$ approaches $P_X$ in this case and $d_B^{PU}(x, 0)$ tends to $d_B(x)$. For $a\to -\infty$ the excess risk tends to 1/4 (risk of $d_B(x)$) as the risk of $d_B^{PU}(x, s)$ tends to 0. 
        \end{example}
    
    \section{Experiment dataset statistics}
    \label{appendix:datasets}
    
        Table \ref{tab:dataset-stats} summarized the properties of all datasets used in the experiment section of the paper.
    
        \begin{table}[h!]
            \centering
            \caption{Dataset statistics}
            \label{tab:dataset-stats}
            \begin{tabular}{lrrr}
                \toprule
                               \textbf{Name} & \textbf{Samples} & \textbf{Features} & \textbf{Class prior} $\bm{\pi}$ \\
                \midrule
                          MNIST 3v5 &    13454 &       784 &              0.53 \\
                          MNIST OvE &    70000 &       784 &              0.51 \\
                     CIFAR CarTruck &    12000 &       512 &              0.50 \\
                CIFAR MachineAnimal &    60000 &       512 &              0.40 \\
                  STL MachineAnimal &    13000 &       512 &              0.40 \\
                       CDC Diabetes &    148458 &       38 &              0.50 \\
                \bottomrule
            \end{tabular}
        \end{table}

    \FloatBarrier

    \section{Balanced accuracy results}
    \label{appendix:balanced_accuracy}

        Tables \ref{tab:U-Balanced_accuracy-Synth+S} through \ref{tab:U-Balanced_accuracy-Real-Compare} correspond to tables \ref{tab:U-Accuracy-Synth+S}-\ref{tab:U-Accuracy-Real-Compare} in the main paper and present U-balanced accuracy for all the performed experiments. This is an important metric, as U-metrics introduce imbalance into the measurements, which might impact accuracy negatively. We can see, however, that the obtained results are close to results presented in the main paper -- while VP-B+S does not have as overwhelming of an advantage, it still clearly is the best method overall.

        \begin{table}[h!]
	\centering \scriptsize \renewcommand{\arraystretch}{1.2}
	\caption{U-Balanced accuracy values -- Method comparison -- Synthetic datasets}
	\label{tab:U-Balanced_accuracy-Synth+S}
\scalebox{0.80}{
	\begin{tabular}{llllll}
		\toprule
		\textbf{c} & \textbf{Method} & \textbf{Synth. 1} & \textbf{Synth. 2} & \textbf{Synth. 3} & \textbf{Synth. SCAR} \\
		\midrule
		\multirow{3}{*}{0.02} & VP+S &  \textbf{61.29 $\pm$ 2.27} &  \textbf{59.43 $\pm$ 2.66} &  \textbf{60.53 $\pm$ 2.30} &  \textbf{63.52 $\pm$ 1.77} \\
		     & VP-B+S &           61.04 $\pm$ 2.33 &           59.32 $\pm$ 2.61 &           60.40 $\pm$ 2.25 &           63.33 $\pm$ 1.73 \\
		     & LBE+S &           49.70 $\pm$ 0.35 &           50.01 $\pm$ 0.35 &           49.93 $\pm$ 0.32 &           49.69 $\pm$ 0.33 \\
		\cline{1-6}
		\multirow{3}{*}{0.10} & VP+S &           67.67 $\pm$ 0.52 &  \textbf{66.72 $\pm$ 0.60} &  \textbf{67.47 $\pm$ 0.61} &  \textbf{68.49 $\pm$ 0.47} \\
		     & VP-B+S &  \textbf{67.71 $\pm$ 0.52} &           66.45 $\pm$ 0.54 &           67.43 $\pm$ 0.58 &           68.26 $\pm$ 0.47 \\
		     & LBE+S &           50.03 $\pm$ 0.33 &           50.83 $\pm$ 0.36 &           50.22 $\pm$ 0.32 &           49.66 $\pm$ 0.37 \\
		\cline{1-6}
		\multirow{3}{*}{0.30} & VP+S &           67.49 $\pm$ 0.58 &           65.39 $\pm$ 0.59 &           66.24 $\pm$ 0.50 &  \textbf{70.33 $\pm$ 0.56} \\
		     & VP-B+S &  \textbf{67.74 $\pm$ 0.56} &  \textbf{65.61 $\pm$ 0.63} &  \textbf{66.35 $\pm$ 0.52} &           70.17 $\pm$ 0.55 \\
		     & LBE+S &           52.86 $\pm$ 0.18 &           52.88 $\pm$ 0.33 &           52.68 $\pm$ 0.22 &           50.00 $\pm$ 0.32 \\
		\cline{1-6}
		\multirow{3}{*}{0.50} & VP+S &  \textbf{64.11 $\pm$ 0.54} &  \textbf{61.68 $\pm$ 0.62} &           63.18 $\pm$ 0.67 &  \textbf{69.30 $\pm$ 0.43} \\
		     & VP-B+S &           64.09 $\pm$ 0.53 &           61.65 $\pm$ 0.61 &  \textbf{63.27 $\pm$ 0.70} &           69.21 $\pm$ 0.44 \\
		     & LBE+S &           55.57 $\pm$ 0.49 &           54.62 $\pm$ 0.47 &           54.96 $\pm$ 0.50 &           56.69 $\pm$ 0.44 \\
		\cline{1-6}
		\multirow{3}{*}{0.70} & VP+S &  \textbf{59.40 $\pm$ 0.78} &           58.33 $\pm$ 0.73 &  \textbf{58.51 $\pm$ 0.65} &           65.79 $\pm$ 0.56 \\
		     & VP-B+S &  \textbf{59.40 $\pm$ 0.79} &  \textbf{58.42 $\pm$ 0.69} &           58.35 $\pm$ 0.60 &           65.49 $\pm$ 0.71 \\
		     & LBE+S &           58.33 $\pm$ 0.65 &           56.61 $\pm$ 0.77 &           56.60 $\pm$ 0.66 &  \textbf{67.87 $\pm$ 0.59} \\
		\cline{1-6}
		\multirow{3}{*}{0.90} & VP+S &           52.19 $\pm$ 0.57 &           52.48 $\pm$ 0.58 &           52.33 $\pm$ 0.51 &           59.52 $\pm$ 0.98 \\
		     & VP-B+S &           52.19 $\pm$ 0.55 &           52.41 $\pm$ 0.57 &           52.48 $\pm$ 0.60 &           58.64 $\pm$ 0.85 \\
		     & LBE+S &  \textbf{57.42 $\pm$ 0.86} &  \textbf{56.47 $\pm$ 0.92} &  \textbf{54.63 $\pm$ 1.01} &  \textbf{71.37 $\pm$ 1.02} \\
		\bottomrule
	\end{tabular}
}
\end{table}

        \begin{table}[h!]
	\centering \scriptsize \renewcommand{\arraystretch}{1.2}
	\caption{U-Balanced accuracy values -- Method comparison -- Real-world datasets}
	\label{tab:U-Balanced_accuracy-Real+S}
\scalebox{0.63}{
	\begin{tabular}{llllllll}
		\toprule
		\textbf{c} & \textbf{Method} & \textbf{MNIST 3v5} & \textbf{MNIST OvE} & \textbf{CIFAR CT} & \textbf{CIFAR MA} & \textbf{STL MA} & \textbf{CDC-Diabetes} \\
		\midrule
		\multirow{3}{*}{0.02} & VP+S &           76.89 $\pm$ 1.14 &           68.32 $\pm$ 1.19 &           87.28 $\pm$ 0.49 &           91.24 $\pm$ 0.21 &           82.13 $\pm$ 0.64 &           50.00 $\pm$ 1.48 \\
		     & VP-B+S &  \textbf{78.37 $\pm$ 1.53} &  \textbf{74.88 $\pm$ 1.51} &  \textbf{92.46 $\pm$ 0.43} &  \textbf{94.03 $\pm$ 0.09} &  \textbf{83.42 $\pm$ 0.69} &  \textbf{51.16 $\pm$ 1.74} \\
		     & LBE+S &           49.95 $\pm$ 0.34 &           50.14 $\pm$ 0.14 &           50.00 $\pm$ 0.40 &           50.16 $\pm$ 0.18 &           50.17 $\pm$ 0.24 &           49.97 $\pm$ 0.20 \\
		\cline{1-8}
		\multirow{3}{*}{0.10} & VP+S &           80.06 $\pm$ 0.61 &           74.69 $\pm$ 1.38 &           91.64 $\pm$ 0.32 &           92.75 $\pm$ 0.24 &           87.19 $\pm$ 0.32 &           57.91 $\pm$ 0.78 \\
		     & VP-B+S &  \textbf{84.29 $\pm$ 0.76} &  \textbf{83.22 $\pm$ 1.23} &  \textbf{93.46 $\pm$ 0.19} &  \textbf{94.28 $\pm$ 0.13} &  \textbf{88.24 $\pm$ 0.30} &  \textbf{60.93 $\pm$ 0.68} \\
		     & LBE+S &           50.25 $\pm$ 0.34 &           49.79 $\pm$ 0.15 &           50.61 $\pm$ 0.38 &           50.34 $\pm$ 0.28 &           50.78 $\pm$ 0.28 &           49.93 $\pm$ 0.19 \\
		\cline{1-8}
		\multirow{3}{*}{0.30} & VP+S &           82.31 $\pm$ 0.67 &           80.74 $\pm$ 0.86 &           93.20 $\pm$ 0.26 &           94.06 $\pm$ 0.10 &  \textbf{88.65 $\pm$ 0.28} &           57.47 $\pm$ 0.84 \\
		     & VP-B+S &  \textbf{86.95 $\pm$ 0.51} &  \textbf{87.99 $\pm$ 0.64} &  \textbf{94.20 $\pm$ 0.16} &  \textbf{94.78 $\pm$ 0.09} &           88.32 $\pm$ 0.35 &  \textbf{61.60 $\pm$ 0.92} \\
		     & LBE+S &           50.77 $\pm$ 0.31 &           49.79 $\pm$ 0.18 &           55.32 $\pm$ 1.30 &           60.51 $\pm$ 4.24 &           55.90 $\pm$ 1.85 &           49.93 $\pm$ 0.17 \\
		\cline{1-8}
		\multirow{3}{*}{0.50} & VP+S &           85.05 $\pm$ 0.70 &           84.64 $\pm$ 0.78 &           94.24 $\pm$ 0.24 &  \textbf{94.52 $\pm$ 0.12} &  \textbf{89.34 $\pm$ 0.47} &           59.37 $\pm$ 0.56 \\
		     & VP-B+S &  \textbf{88.49 $\pm$ 0.83} &  \textbf{89.86 $\pm$ 0.59} &  \textbf{94.73 $\pm$ 0.24} &           94.43 $\pm$ 0.15 &           88.60 $\pm$ 0.53 &  \textbf{62.98 $\pm$ 0.41} \\
		     & LBE+S &           51.80 $\pm$ 0.34 &           52.90 $\pm$ 1.30 &           72.38 $\pm$ 2.72 &           73.44 $\pm$ 2.77 &           70.34 $\pm$ 2.07 &           62.39 $\pm$ 1.83 \\
		\cline{1-8}
		\multirow{3}{*}{0.70} & VP+S &           88.28 $\pm$ 0.67 &           89.25 $\pm$ 0.54 &           94.35 $\pm$ 0.32 &  \textbf{94.23 $\pm$ 0.17} &  \textbf{88.63 $\pm$ 0.54} &           60.07 $\pm$ 0.73 \\
		     & VP-B+S &  \textbf{90.09 $\pm$ 0.64} &  \textbf{91.30 $\pm$ 0.55} &  \textbf{94.66 $\pm$ 0.27} &           93.51 $\pm$ 0.26 &           87.97 $\pm$ 0.51 &           62.31 $\pm$ 0.62 \\
		     & LBE+S &           54.66 $\pm$ 0.65 &           60.61 $\pm$ 1.00 &           91.25 $\pm$ 1.47 &           90.31 $\pm$ 1.01 &           83.02 $\pm$ 1.48 &  \textbf{69.67 $\pm$ 1.02} \\
		\cline{1-8}
		\multirow{3}{*}{0.90} & VP+S &  \textbf{87.14 $\pm$ 0.74} &  \textbf{92.30 $\pm$ 0.40} &           91.97 $\pm$ 0.55 &           92.77 $\pm$ 0.45 &           82.75 $\pm$ 1.47 &           63.01 $\pm$ 0.71 \\
		     & VP-B+S &           86.15 $\pm$ 0.70 &           91.84 $\pm$ 0.60 &           92.00 $\pm$ 0.55 &           89.65 $\pm$ 0.34 &           82.43 $\pm$ 1.45 &           58.52 $\pm$ 0.53 \\
		     & LBE+S &           67.77 $\pm$ 0.98 &           83.49 $\pm$ 0.79 &  \textbf{93.50 $\pm$ 0.46} &  \textbf{93.68 $\pm$ 0.46} &  \textbf{88.35 $\pm$ 0.70} &  \textbf{72.88 $\pm$ 0.38} \\
		\bottomrule
	\end{tabular}
}
\end{table}

        \begin{table}[h!]
	\centering \scriptsize \renewcommand{\arraystretch}{1.2}
	\caption{U-Balanced accuracy values -- Decision rule comparison -- Synthetic datasets}
	\label{tab:U-Balanced_accuracy-Synth-Compare}
\scalebox{0.80}{
	\begin{tabular}{llllll}
		\toprule
		\textbf{c} & \textbf{Method} & \textbf{Synth. 1} & \textbf{Synth. 2} & \textbf{Synth. 3} & \textbf{Synth. SCAR} \\
		\midrule
		\multirow{6}{*}{0.02} & S-Prophet &           73.29 $\pm$ 0.35 &           73.25 $\pm$ 0.36 &           71.37 $\pm$ 0.35 &           73.48 $\pm$ 0.35 \\
		     & Y-Prophet &           73.32 $\pm$ 0.36 &           73.25 $\pm$ 0.36 &           71.40 $\pm$ 0.36 &           73.51 $\pm$ 0.35 \\
		     & VP-B &           60.94 $\pm$ 2.39 &  \textbf{59.42 $\pm$ 2.55} &           60.14 $\pm$ 2.31 &  \textbf{63.33 $\pm$ 1.68} \\
		     & VP-B+S &  \textbf{61.04 $\pm$ 2.33} &           59.32 $\pm$ 2.61 &  \textbf{60.40 $\pm$ 2.25} &  \textbf{63.33 $\pm$ 1.73} \\
		     & VP-B+S + true s(x) &           60.81 $\pm$ 2.36 &           59.16 $\pm$ 2.68 &           60.18 $\pm$ 2.29 &           63.23 $\pm$ 1.70 \\
		     & VP-B+S + true y(x) &           73.29 $\pm$ 0.37 &           73.22 $\pm$ 0.36 &           71.43 $\pm$ 0.36 &           73.46 $\pm$ 0.34 \\
		\cline{1-6}
		\multirow{6}{*}{0.10} & S-Prophet &           72.52 $\pm$ 0.31 &           72.06 $\pm$ 0.35 &           70.49 $\pm$ 0.31 &           73.61 $\pm$ 0.35 \\
		     & Y-Prophet &           72.60 $\pm$ 0.35 &           72.13 $\pm$ 0.38 &           70.65 $\pm$ 0.37 &           73.69 $\pm$ 0.33 \\
		     & VP-B &  \textbf{68.02 $\pm$ 0.46} &  \textbf{66.65 $\pm$ 0.57} &  \textbf{67.62 $\pm$ 0.59} &           68.16 $\pm$ 0.38 \\
		     & VP-B+S &           67.71 $\pm$ 0.52 &           66.45 $\pm$ 0.54 &           67.43 $\pm$ 0.58 &  \textbf{68.26 $\pm$ 0.47} \\
		     & VP-B+S + true s(x) &           67.80 $\pm$ 0.52 &           66.57 $\pm$ 0.49 &           67.30 $\pm$ 0.62 &           68.20 $\pm$ 0.46 \\
		     & VP-B+S + true y(x) &           72.53 $\pm$ 0.35 &           71.75 $\pm$ 0.41 &           70.42 $\pm$ 0.36 &           73.55 $\pm$ 0.33 \\
		\cline{1-6}
		\multirow{6}{*}{0.30} & S-Prophet &           69.91 $\pm$ 0.47 &           69.02 $\pm$ 0.52 &           67.65 $\pm$ 0.42 &           72.69 $\pm$ 0.48 \\
		     & Y-Prophet &           70.57 $\pm$ 0.43 &           69.37 $\pm$ 0.42 &           68.49 $\pm$ 0.39 &           73.60 $\pm$ 0.35 \\
		     & VP-B &  \textbf{68.08 $\pm$ 0.52} &  \textbf{65.85 $\pm$ 0.58} &  \textbf{67.19 $\pm$ 0.52} &  \textbf{70.53 $\pm$ 0.49} \\
		     & VP-B+S &           67.74 $\pm$ 0.56 &           65.61 $\pm$ 0.63 &           66.35 $\pm$ 0.52 &           70.17 $\pm$ 0.55 \\
		     & VP-B+S + true s(x) &           67.67 $\pm$ 0.57 &           65.70 $\pm$ 0.64 &           66.48 $\pm$ 0.55 &           70.14 $\pm$ 0.51 \\
		     & VP-B+S + true y(x) &           69.30 $\pm$ 0.47 &           68.33 $\pm$ 0.52 &           67.14 $\pm$ 0.43 &           72.61 $\pm$ 0.50 \\
		\cline{1-6}
		\multirow{6}{*}{0.50} & S-Prophet &           65.33 $\pm$ 0.70 &           63.89 $\pm$ 0.56 &           63.25 $\pm$ 0.71 &           71.13 $\pm$ 0.62 \\
		     & Y-Prophet &           68.13 $\pm$ 0.49 &           66.56 $\pm$ 0.48 &           66.02 $\pm$ 0.50 &           73.18 $\pm$ 0.36 \\
		     & VP-B &  \textbf{64.95 $\pm$ 0.45} &  \textbf{62.65 $\pm$ 0.56} &  \textbf{64.12 $\pm$ 0.70} &  \textbf{70.31 $\pm$ 0.49} \\
		     & VP-B+S &           64.09 $\pm$ 0.53 &           61.65 $\pm$ 0.61 &           63.27 $\pm$ 0.70 &           69.21 $\pm$ 0.44 \\
		     & VP-B+S + true s(x) &           64.08 $\pm$ 0.48 &           61.58 $\pm$ 0.60 &           63.15 $\pm$ 0.67 &           69.30 $\pm$ 0.46 \\
		     & VP-B+S + true y(x) &           64.44 $\pm$ 0.62 &           62.68 $\pm$ 0.60 &           62.03 $\pm$ 0.82 &           70.47 $\pm$ 0.69 \\
		\cline{1-6}
		\multirow{6}{*}{0.70} & S-Prophet &           57.88 $\pm$ 0.51 &           56.62 $\pm$ 0.38 &           56.71 $\pm$ 0.61 &           66.39 $\pm$ 0.48 \\
		     & Y-Prophet &           64.64 $\pm$ 0.70 &           63.34 $\pm$ 0.70 &           62.71 $\pm$ 0.73 &           73.39 $\pm$ 0.41 \\
		     & VP-B &  \textbf{61.12 $\pm$ 0.65} &  \textbf{59.53 $\pm$ 0.68} &  \textbf{59.84 $\pm$ 0.68} &  \textbf{68.95 $\pm$ 0.68} \\
		     & VP-B+S &           59.40 $\pm$ 0.79 &           58.42 $\pm$ 0.69 &           58.35 $\pm$ 0.60 &           65.49 $\pm$ 0.71 \\
		     & VP-B+S + true s(x) &           59.33 $\pm$ 0.61 &           58.45 $\pm$ 0.61 &           58.31 $\pm$ 0.70 &           65.01 $\pm$ 0.78 \\
		     & VP-B+S + true y(x) &           57.07 $\pm$ 0.73 &           56.58 $\pm$ 0.52 &           56.44 $\pm$ 0.64 &           65.76 $\pm$ 0.39 \\
		\cline{1-6}
		\multirow{6}{*}{0.90} & S-Prophet &           50.38 $\pm$ 0.19 &           50.35 $\pm$ 0.21 &           50.47 $\pm$ 0.16 &           55.62 $\pm$ 0.55 \\
		     & Y-Prophet &           61.12 $\pm$ 1.16 &           60.29 $\pm$ 1.07 &           58.82 $\pm$ 1.08 &           72.72 $\pm$ 1.00 \\
		     & VP-B &  \textbf{58.85 $\pm$ 0.62} &  \textbf{57.12 $\pm$ 0.69} &  \textbf{55.19 $\pm$ 0.99} &  \textbf{71.61 $\pm$ 0.91} \\
		     & VP-B+S &           52.19 $\pm$ 0.55 &           52.41 $\pm$ 0.57 &           52.48 $\pm$ 0.60 &           58.64 $\pm$ 0.85 \\
		     & VP-B+S + true s(x) &           52.99 $\pm$ 0.40 &           52.90 $\pm$ 0.41 &           51.86 $\pm$ 0.67 &           60.43 $\pm$ 0.93 \\
		     & VP-B+S + true y(x) &           50.94 $\pm$ 0.53 &           51.03 $\pm$ 0.41 &           52.25 $\pm$ 0.50 &           54.80 $\pm$ 0.55 \\
		\bottomrule
	\end{tabular}
}
\end{table}

        \begin{table}[h!]
	\centering \scriptsize \renewcommand{\arraystretch}{1.2}
	\caption{U-Balanced accuracy values -- Decision rule comparison -- Real-world datasets}
	\label{tab:U-Balanced_accuracy-Real-Compare}
\scalebox{0.63}{
	\begin{tabular}{llllllll}
		\toprule
		\textbf{c} & \textbf{Method} & \textbf{MNIST 3v5} & \textbf{MNIST OvE} & \textbf{CIFAR CT} & \textbf{CIFAR MA} & \textbf{STL MA} & \textbf{CDC-Diabetes} \\
		\midrule
		\multirow{2}{*}{0.02} & VP-B &           78.32 $\pm$ 1.45 &           74.49 $\pm$ 1.49 &           92.42 $\pm$ 0.41 &           93.91 $\pm$ 0.09 &  \textbf{83.46 $\pm$ 0.68} &           51.12 $\pm$ 1.76 \\
		     & VP-B+S &  \textbf{78.37 $\pm$ 1.53} &  \textbf{74.88 $\pm$ 1.51} &  \textbf{92.46 $\pm$ 0.43} &  \textbf{94.03 $\pm$ 0.09} &           83.42 $\pm$ 0.69 &  \textbf{51.16 $\pm$ 1.74} \\
		\cline{1-8}
		\multirow{2}{*}{0.10} & VP-B &           84.10 $\pm$ 0.65 &           82.80 $\pm$ 1.28 &           93.35 $\pm$ 0.19 &  \textbf{94.28 $\pm$ 0.12} &  \textbf{88.27 $\pm$ 0.29} &  \textbf{61.38 $\pm$ 0.77} \\
		     & VP-B+S &  \textbf{84.29 $\pm$ 0.76} &  \textbf{83.22 $\pm$ 1.23} &  \textbf{93.46 $\pm$ 0.19} &  \textbf{94.28 $\pm$ 0.13} &           88.24 $\pm$ 0.30 &           60.93 $\pm$ 0.68 \\
		\cline{1-8}
		\multirow{2}{*}{0.30} & VP-B &           86.75 $\pm$ 0.55 &  \textbf{88.04 $\pm$ 0.65} &           94.12 $\pm$ 0.18 &  \textbf{94.85 $\pm$ 0.08} &  \textbf{89.11 $\pm$ 0.28} &  \textbf{63.78 $\pm$ 0.78} \\
		     & VP-B+S &  \textbf{86.95 $\pm$ 0.51} &           87.99 $\pm$ 0.64 &  \textbf{94.20 $\pm$ 0.16} &           94.78 $\pm$ 0.09 &           88.32 $\pm$ 0.35 &           61.60 $\pm$ 0.92 \\
		\cline{1-8}
		\multirow{2}{*}{0.50} & VP-B &           88.33 $\pm$ 0.88 &  \textbf{90.09 $\pm$ 0.57} &           94.69 $\pm$ 0.25 &  \textbf{94.72 $\pm$ 0.10} &  \textbf{89.81 $\pm$ 0.34} &  \textbf{67.36 $\pm$ 0.25} \\
		     & VP-B+S &  \textbf{88.49 $\pm$ 0.83} &           89.86 $\pm$ 0.59 &  \textbf{94.73 $\pm$ 0.24} &           94.43 $\pm$ 0.15 &           88.60 $\pm$ 0.53 &           62.98 $\pm$ 0.41 \\
		\cline{1-8}
		\multirow{2}{*}{0.70} & VP-B &  \textbf{90.23 $\pm$ 0.66} &  \textbf{92.15 $\pm$ 0.47} &  \textbf{94.81 $\pm$ 0.26} &  \textbf{94.65 $\pm$ 0.15} &  \textbf{90.82 $\pm$ 0.36} &  \textbf{68.93 $\pm$ 0.40} \\
		     & VP-B+S &           90.09 $\pm$ 0.64 &           91.30 $\pm$ 0.55 &           94.66 $\pm$ 0.27 &           93.51 $\pm$ 0.26 &           87.97 $\pm$ 0.51 &           62.31 $\pm$ 0.62 \\
		\cline{1-8}
		\multirow{2}{*}{0.90} & VP-B &  \textbf{87.31 $\pm$ 0.72} &  \textbf{93.55 $\pm$ 0.45} &  \textbf{94.07 $\pm$ 0.33} &  \textbf{94.89 $\pm$ 0.19} &  \textbf{90.03 $\pm$ 0.73} &  \textbf{71.38 $\pm$ 0.37} \\
		     & VP-B+S &           86.15 $\pm$ 0.70 &           91.84 $\pm$ 0.60 &           92.00 $\pm$ 0.55 &           89.65 $\pm$ 0.34 &           82.43 $\pm$ 1.45 &           58.52 $\pm$ 0.53 \\
		\bottomrule
	\end{tabular}
}
\end{table}

    \FloatBarrier
        
\end{appendices}


\end{document}